\documentclass{elsarticle}
\usepackage{graphicx}
\usepackage[utf8]{inputenc}
\usepackage{epstopdf}
\usepackage[dvipsnames]{xcolor}
\usepackage{enumitem}
\usepackage{times}
\usepackage[OT2,T2A,T1]{fontenc}
\usepackage[main=english,russian]{babel}
\usepackage{lmodern}
\usepackage{amsmath}
\usepackage{amsfonts}
\usepackage{amssymb}
\usepackage{geometry}

\usepackage{subfloat} 
\usepackage{caption}
\usepackage{subcaption} 

\usepackage{comment}

\newgeometry{lmargin=1.75cm, rmargin=1.75cm, tmargin=2cm, bmargin=2cm, ignorehead, ignorefoot}

\newenvironment{packed_itemize}{
\begin{itemize}
  \setlength{\itemsep}{0.25pt}
  \setlength{\parskip}{0pt}
  \setlength{\parsep}{0pt}
}{\end{itemize}}

\begin{document}
\begin{frontmatter}

\title{Universal versus system-specific features of punctuation usage patterns in~major Western~languages}

\author{Tomasz Stanisz$^1$, Stanisław Drożdż$^{1,2}$, Jarosław Kwapień$^{1}$}

\address{$^1$Complex Systems Theory Department, Institute of Nuclear Physics, Polish Academy of Sciences, ul. Radzikowskiego 152, 31-342 Kraków, Poland \\ $^2$Faculty of Computer Science and Telecommunications, Cracow University of Technology, ul. Warszawska 24, Kraków 31-155, Poland}

\begin{abstract}

The celebrated proverb that ``speech is silver, silence is golden'' has a long multinational history and multiple specific meanings. In written texts punctuation can in fact be considered one of its manifestations. Indeed, the virtue of effectively speaking and writing involves - often decisively - the capacity to apply the properly placed breaks. In the present study, based on a large corpus of world-famous and representative literary texts in seven major Western languages, it is shown that the distribution of intervals between consecutive punctuation marks in almost all texts can universally be characterised by only two parameters of the discrete Weibull distribution which can be given an intuitive interpretation in terms of the so-called hazard function. The values of these two parameters tend to be language-specific, however, and even appear to navigate translations. The properties of the computed hazard functions indicate that among the studied languages, English turns out to be the least constrained by the necessity to place a consecutive punctuation mark to partition a sequence of words. This may suggest that when compared to other studied languages, English is more flexible, in the sense of allowing longer uninterrupted sequences of words. Spanish reveals similar tendency to only a bit lesser extent.

\end{abstract}

\begin{keyword}
Natural language \sep Punctuation \sep Survival analysis \sep Complexity \sep Quantitative linguistics
\end{keyword}

\end{frontmatter}

\section{Introduction}

The faculty of language~\cite{Hauser2002} establishes the organization of societies~\cite{Harris1987,Pinker1994,Pagel2012,Gelman2017} and is commonly considered the principal trait of humanity~\cite{Bloom2001,Sole2010,Pagel2017}. Yet its evolutionary emergence~\cite{Darwin1871,Bolhuis2014}, involving highly convoluted cognition and computation related processes~\cite{Nowak2002,Goldin-Meadow2013,Bouchard2013,Rabinovich2020} in the brain, still remains a subject of controversies~\cite{Hauser2002,Pinker1990,Christiansen2008,Christiansen2016}. The externalization of these processes in terms of speech~\cite{Hockett1960}, consecutively mapped by humans into writing~\cite{Algeo2010} - where punctuation plays a pivotal role - provides the most straightforward and quantifiable clue to the design principles of natural language, involving both neuronal and anatomical processes. This especially applies to its efficiency and expressiveness in communication. As one related element, the arrangement of full stops in a written text determines the resulting sentence length distribution while the arrangement of all the punctuation reveals characteristics of the basic constituents in the syntactic organization of a sentence~\cite{Chomsky1965}.

Even though the usage of punctuation is crucial for a faithful mapping of the spoken language into writing~\cite{Say1997,Iyer2001,Baron2001}, the history of its development is relatively young and recent~\cite{Parkes1993,Nunberg2002,Truss2003} as compared to the time horizons involved in the development of modern languages~\cite{Gray2003}. In its developed, contemporary Western form, by partitioning texts through several punctuation marks, it serves clarifying meaning, expressing the emotional content, facilitating perception, memorization and the cognitive processing, as dictated by perceptuo-motor factors, cognitive limitations, and pragmatics, both of a writer and of a reader. Interestingly, punctuation marks, accompanying words as the ''building blocks'' of written texts, share some specific quantitative characteristics with words, despite being clearly different in their nature and role in language. Recent statistical analysis shows~\cite{Kulig2017} that punctuation marks obey the same Zipf's law~\cite{Zipf1949} as words. Even more, their inclusion in word frequency analysis acts towards restoring the Zipf's power law from the more flat Zipf–Mandelbrot~\cite{Mandelbrot1966} behaviour in the initial region of the most frequent items and thus improves an overall quality of scaling.  A related practical result is that treating punctuation marks on the same terms as words appears to have a significant impact on the accuracy of stylometric analyses~\cite{Stanisz2019}.

\section{Datasets and Methods}

\subsection{Literary corpora studied}

The main set of studied texts consists of 240 literary works written in prose in 7 languages: English (denoted by EN in figures), German (DE), French (FR), Italian (IT), Spanish (ES), Polish (PL) and Russian (RU). In each language there are more than 30 texts, summing up to corpora of at least 3 million words in length. Each book contains no fewer than 1500 word sequences separated by punctuation marks. In terms of sentences, the lengths of the books are also typically on the order of thousands; there are only 8 texts shorter than 1000 sentences, with the shortest one having above 350 sentences. An additional book set consists of translated texts. This set is constructed as follows. In each of the studied languages, 2 books from the main set are selected. Then their translations into all the remaining languages are added into the set. One thus gets a collection of 14 books, each in 7 language versions. Two translations are missing in the set (due to unavailability), and thus 96 texts are obtained in total. Full list of all the books analysed is given in~\ref{sect::appendix-c}. The main criteria for the choice of texts are availability (the majority are in the public domain) and popularity - many of them are famous pieces of world literature having a significant impact on culture. The literature in each of the 7 languages included was awarded no fewer than 5 Nobel Prizes and each of them is spoken by more than 40 million people worldwide. 

The texts are subjected to preprocessing, which includes removing foreword, annotations, additional information from the publisher, and punctuation marks that are not related to the structure of a sentence, for example hyphens, full stops used after abbreviations, dashes placed at the beginning of quotations in some languages, and inverted interrogation and exclamation marks in Spanish. Each text is transformed into a series $k_i$ representing the lengths of intervals $i$ between consecutive punctuation marks, measured by the number of words (disregarding $k_i=0$, that is the cases where two punctuation marks are placed next to each other). Three variants of what is taken as punctuation are quantitatively considered: (i) punctuation marks indicating the ends of sentences, (ii) sentence-ending punctuation marks together with all the commas and (iii) all punctuation marks (after preprocessing these belong to the following list: full stop, question mark, exclamation mark, ellipsis, comma, dash, colon, semicolon, left and right bracket).

\subsection{Discrete Weibull distribution}

Since the usage of punctuation is inherently related to the life processes, and thus also to their limitations and even exhaustion, some of their exposure characteristics are expected to reflect this fact and thus to be suitable to the survival analysis~\cite{Miller1997}. Indeed, the distribution of punctuation recurrence patterns can be considered the main factor reflecting the 'lifetime' of an uninterrupted sequence of words. A candidate functional form to describe the distribution of inter-punctuation distances $k$ within a written text is a Weibull-like distribution~\cite{Frechet1927,Weibull1951} as it often appears to properly model the processes related to survival. 
%
%

To understand the motivation behind introducing discrete Weibull distribution as a model of the distribution of distances between consecutive punctuation marks, one can consider the following line of reasoning. 

Suppose that putting punctuation marks into the text is governed by a random process: after each word written, the writer places a punctuation mark with probability $p$, and puts no mark with probability $1-p$. Here no distinction between different types of punctuation marks is made; this is motivated by the fact that many punctuation marks, despite being clearly different, can be considered as serving roughly the same purpose - to break text into parts, in the way imposed by grammatical and logical constraints. Each decision whether to put a punctuation mark or not is a Bernoulli trial with $p$ being the probability of success. The length of the interval between successive punctuation marks (the number of words between them) is the number $k$ of trials required to obtain the first success, after the last one observed ($k=1,2,3,...$). The number of trials until the first success in Bernoulli process  follows the geometric distribution. 

This distribution, just as any other, can be fully characterised by its cumulative distribution function $F(k)$, or equivalently, by giving the complementary cumulative distribution function $1-F(k)$, also called the \textit{survival function}, which describes the probability that the random variable takes on a value greater than $k$. The survival function of geometric distribution is:
\begin{equation}
1-F(k)=\left( 1-p \right)^k.
\end{equation}
However, it is reasonable to anticipate that the probability of the punctuation mark occurrence after a word is not constant over the text, and therefore a distribution more general than the geometric distribution is required. One way of generalising the geometric distribution is introducing an exponent, $\beta>0$, into its survival function:
\begin{equation} \label{eq::DW.CCDF}
1-F(k)=\left( 1-p \right)^{k^\beta}.
\end{equation}
A distribution with survival function of the form given by Eq.~\ref{eq::DW.CCDF} is called the discrete Weibull distribution~\cite{Nakagawa1975}. Its parameter $\beta$ determines the deviation from the geometric distribution, which is recovered for $\beta=1$. It describes how the probability of obtaining a success depends on how many trials have been performed since the last success observed. This dependence can be characterized by the so-called \textit{hazard function} $h(k)$. It can be defined as the conditional probability that a success occurs on the $k$-th trial, given that it has not occurred in the preceding $k\!-\!1$ trials. With $P(k)$ denoting the probability mass function, the hazard function of the discrete Weibull distribution is given by:
\begin{equation}\label{eq::hazard.Weibull}
h(k) = \frac{P(k)}{1-F(k-1)} = \; 1-\left(1-p\right)^{k^{\beta} - (k-1)^{\beta}}.
\end{equation}
For $\beta < 1$, the hazard function is a decreasing function - the probability of observing an event of interest becomes smaller as the waiting time gets longer. For $\beta > 1$, it is increasing with time. For $\beta = 1$, the hazard function is a constant as for the geometric, \textit{memoryless} distribution. The parameter $p$ thus reflects the probability of observing an event occurrence already in the first trial. Exemplary plots of the discrete Weibull distribution and the corresponding hazard rates are shown in Fig.~\ref{fig::two.books}(d).

\subsection{Detrended Fluctuation Analysis (DFA)}

The above-presented mechanism of inserting punctuation into a text attempts to grasp the properties related solely to the probability distribution describing the distances between consecutive punctuation marks. It does not specify whether the processes generating word sequences between consecutive punctuation marks in different places in a text are mutually dependent or not. To investigate the possible presence of such dependencies in a text, one can study the correlations in a time series constructed from the lengths of word sequences between consecutive punctuation marks. In this work, such correlations are characterized with the use of the Hurst exponent~\cite{Hurst1951} - a tool commonly utilized in the description of long-range correlations in time series analysis. The value of the Hurst exponent $H$ provides information about the strength and the character of the correlations in a time series, by measuring how fluctuations of the series behave in various time scales. For $H=1/2$ the time series is uncorrelated. For $H<1/2$ the series is \textit{antipersistent}, that is, its values in consecutive time steps are negatively correlated; for $H>1/2$ it is \textit{persistent}, which means that its consecutive values are correlated positively; in this case, the series is said to have long-range correlations. At present, the detrended fluctuation analysis (DFA)~\cite{Peng1994} is considered the most numerically stable method of computing the Hurst exponent from a real-valued time series; the details of the method are given in \ref{sect::appendix-b}.

\section{Results}

\subsection{Two books}

An explicit demonstration of how effectively the Weibull functional form performs in a typical text is illustrated using the examples of the two frequently addressed texts in the context of quantitative studies. These are {\it Alice's Adventures in Wonderland} by Lewis Carroll~\cite{Kello2010,Ausloos2012,Bian2016} and {\it David Copperfield} by Charles Dickens~\cite{Newman2006,Goh2008,FerreriCancho2010}. The corresponding empirical distributions are fitted with the discrete Weibull distribution by finding numerically the maximum likelihood estimates of $p$ and $\beta$ parameters. Three cases are considered: (i) full stops, (ii) full stops and commas, (iii) all punctuation marks.


\begin{figure}
\centering
\includegraphics[width=1.0\linewidth]{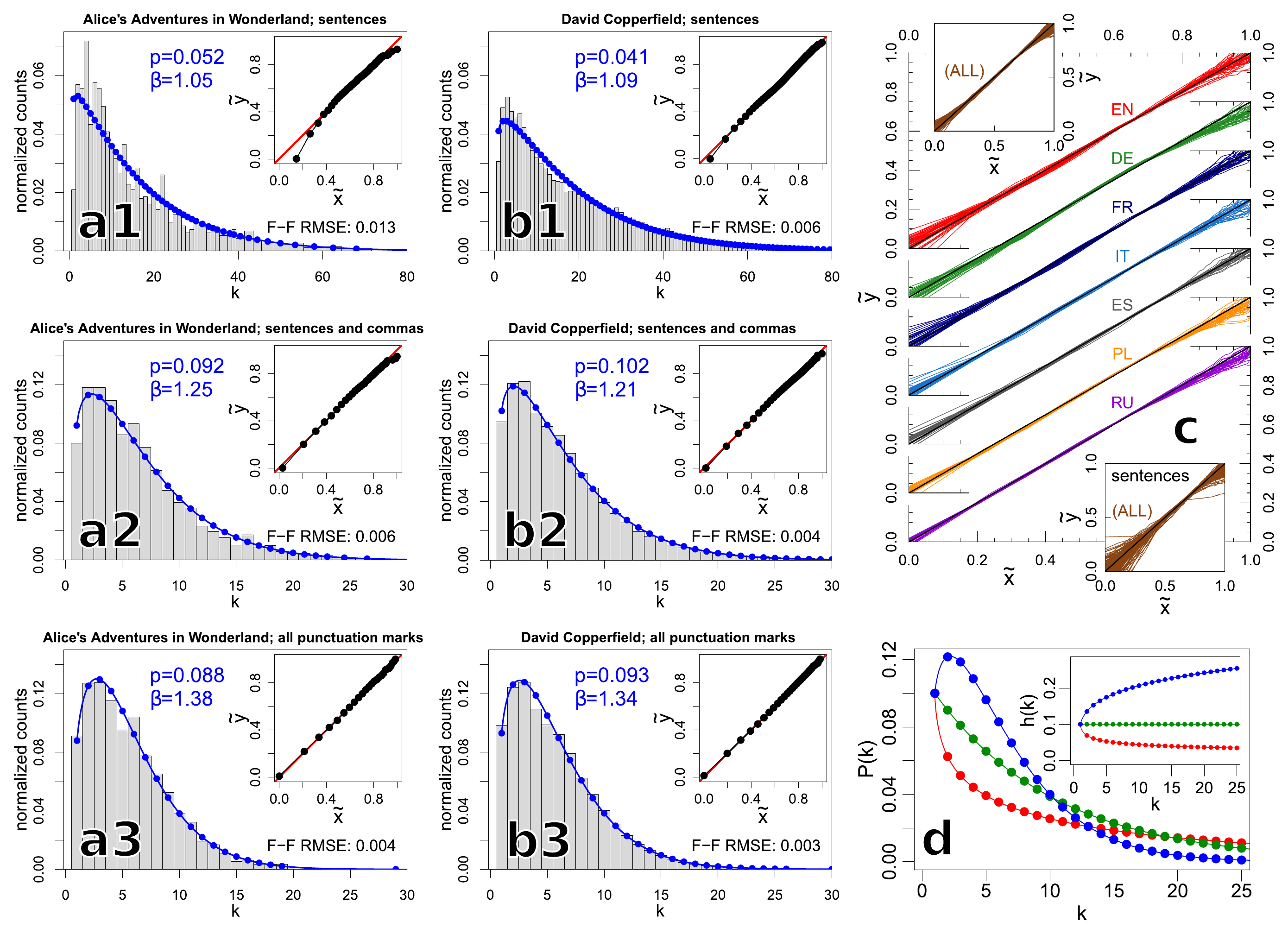}
\caption{Histograms of the distances $k$ between punctuation marks (the distances are measured by the number of words between consecutive punctuation marks), for \textit{Alice's Adventures in Wonderland} by Lewis Carroll (a) and \textit{David Copperfield} by Charles Dickens (b) are considered in the three following cases: In (a1) and (b1) all the sentence-ending marks are included, in (a2) and (b2) the commas are also included, in (a3) and (b3) all the punctuation marks are taken into account. 
The points on the blue curve represent the discrete Weibull distributions fitted to the data (note that the distributions are discrete, and the curves are plotted only as a guide to the eye). In each of the charts, the inset presents the rescaled Weibull plots ($\widetilde{x}, \widetilde{y}$) (defined in~\ref{sect::appendix-a}), which allow to collectively assess the quality of fit to different discrete Weibull distributions; a straight line segment from $(0,0)$ to $(1,1)$ (red) corresponds to a perfect agreement between the empirical distribution and \textit{some} discrete Weibull distribution.
The parameters $p$ and $\beta$ of the fitted distribution and the value of the root mean squared error between the empirical and the fitted cumulative distribution function (F-F RMSE) are given next to each inset. (c) The rescaled Weibull plots of the distributions of distances between consecutive punctuation marks, for 240 books in 7 languages, plotted separately for each language. Each curve represents one book. The inset in the top left corner contains all the plots presented together (without the partition into individual languages). In the bottom right inset, the rescaled Weibull plots of the distributions of sentence lengths are presented, for all the texts in all the studied languages. (d) Exemplary probability mass functions $P(k)$ of discrete Weibull distributions and of the corresponding hazard functions $h(k)$. The parameter $p$ is equal to 0.1 for all three distributions; the values of $\beta$ are: $\beta=0.75$ (red), $\beta=1$ (green; this is the exponential distribution), and $\beta=1.25$ (blue).}
\label{fig::two.books}
\end{figure}

As the juxtaposition of these cases in the corresponding panels (a) and (b) of Fig.~\ref{fig::two.books} shows, such a representation for these texts appears particularly appropriate in terms of accuracy when all the punctuation marks are merged and considered equivalent - thus case (iii). The quality of such a fit somewhat worsens when the sentence-ending marks and commas (case (ii)) are taken into account and the remaining punctuation marks are ignored. In the extreme of case (i), when only the partition of text into sentences is considered, the quality of an analogous fit further degrades such that even its validity can here be questioned. In fact, in this last case the fitted parameter $\beta$ becomes closest to unity which thus signals that the process of the entire sentence length formation is nearer to the uncorrelated geometric distribution. In some cases (e.g. the stream-of-consciousness narrative~\cite{Drozdz2016}) the sentence formation may even depart towards a more 'bursty' dynamics with $\beta < 1$, similarly as in the recurrence patterns of some words~\cite{Altmann2009}. 

The more basic sentence constituents such as clauses, phrases or even single words, as highlighted by the recurrence patterns involving all the punctuation, already involve stronger correlations towards reducing the distribution tails and to make the Weibull distribution (here $\beta$ = 1.38 and 1.34 correspondingly) very efficient in reproducing essentially all its relevant features. It is also interesting to notice that adding only commas (which are one of two most frequent punctuation marks, the other being full stops) to sentence-ending punctuation marks and ignoring all the remaining ones, thus case (ii), still reduces the values of $\beta$ quite substantially and somewhat worsens the quality of fit in terms of the Weibull distribution. This may be taken as an indication that consistency arguments in this kind of study require merging all the punctuation marks.
 
According to Eq.~(\ref{eq::hazard.Weibull}) the parameter $p$ denotes the probability that the next punctuation mark follows the previous one already after one single word, thus for $k=1$. When only the sentence-ending marks are taken into account this probability is expected to assume the lowest values and the corresponding numbers listed in Fig.~\ref{fig::two.books} confirm this. Including in addition the commas (case (ii)) significantly increases this probability which is also anticipated. However, including all the remaining punctuation (case (iii)) somewhat lowers $p$, which signals that it introduces an element of repulsion.

\subsection{Corpora in seven languages}

The above observations appear largely universal as they apply through all the mainstream texts in seven major Western languages considered in the present study: English, French, German, Italian, Polish, Russian, and Spanish. The distributions of the inter-punctuation distances, when all the punctuation marks are taken into account, fit well with the Weibull distribution. For the cases corresponding to (ii), and especially to (i), the quality of such a fit systematically breaks down. For all the 240 books in the two extreme cases of (i) and (iii) this is illustrated in Fig.~\ref{fig::two.books}(c) in terms of the rescaled Weibull plots (see~\ref{sect::appendix-a}). In such a representation an ideal fit is reflected by the diagonal straight line segment. Clearly, in case (iii) all the lines form a narrow band along such a diagonal, thus indicating a good quality fit, while the dispersion becomes significantly larger in case (i).


\begin{figure*}
\centering
\includegraphics[width=1.0\linewidth]{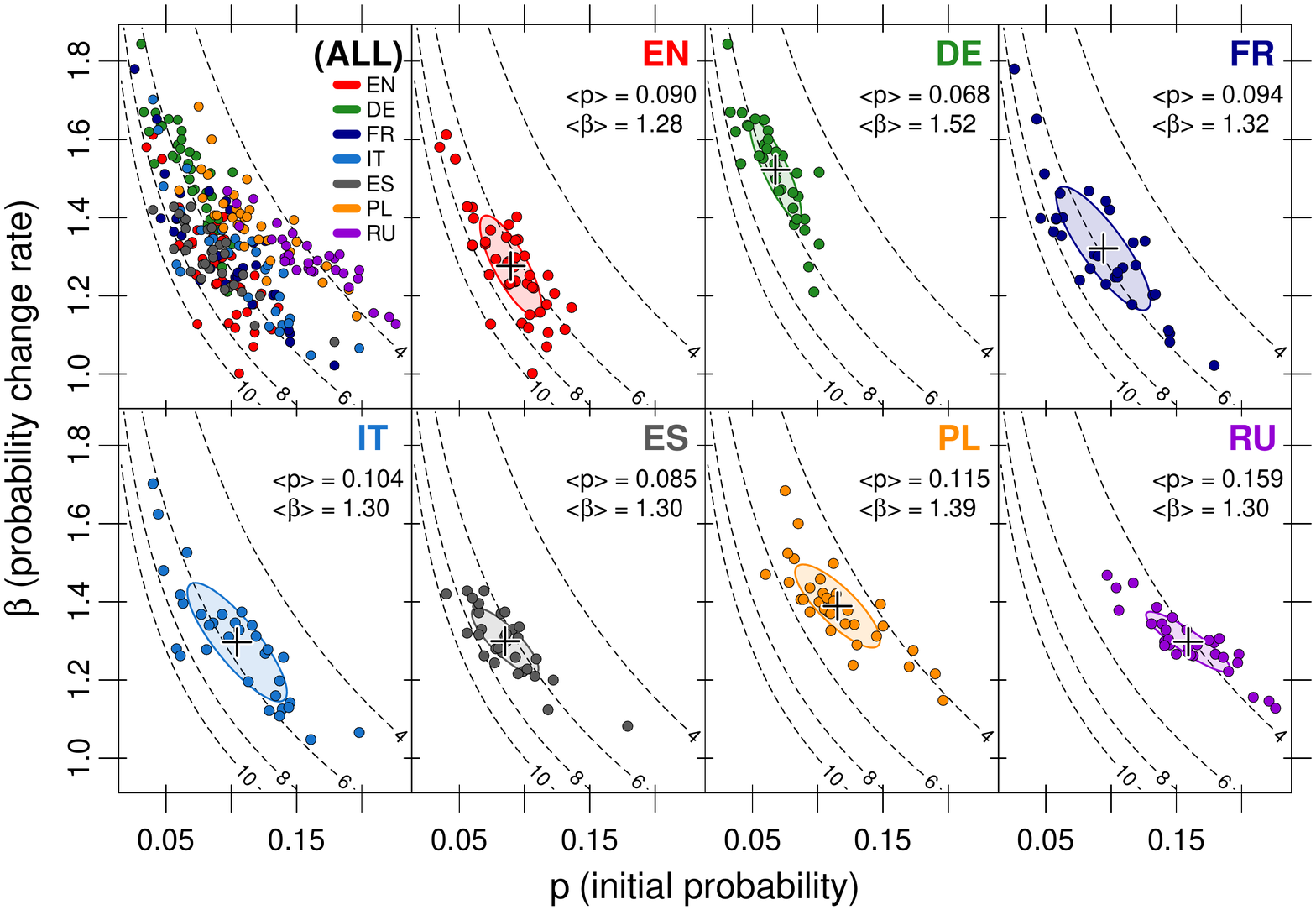}
\caption{The parameters $p$ and $\beta$ of the discrete Weibull distributions fitted to the distances between punctuation marks. The upper left chart (ALL) pertains to all the studied languages collectively, the remaining ones present results for the individual languages. In each plot, a text is represented by a point $(p,\beta)$. All the plots are in the same scale. The dashed lines are isolines of constant expected value of the discrete Weibull distribution - all distributions with $(p,\beta)$ along one such line have the same expected value. In each plot pertaining to a single language, the quantities $\langle p\rangle$ and $\langle\beta\rangle$, the average values of $p$ and $\beta$, are given, and the centroid of the point cloud, $\left(\langle p\rangle,\langle\beta\rangle\right)$ is marked by ''+''. The ellipses characterize the distributions of points - the semi-axes of each ellipse are the principal components of the point set in the given language. The major semi-axis of the ellipse gives the direction of the greatest variance and its length is the square root of that variance. The length of the minor semi-axis is the square root of the variance in the perpendicular direction.}
\label{fig::scatterplots.original}
\end{figure*}

The specific pairs of parameters $\beta$ and $p$ corresponding to the best fits turn out, however, to be significantly varying for different texts as the global scatter plot in the upper left corner of Fig.~\ref{fig::scatterplots.original} highlights it. A closer inspection of the distribution of points in this panel reveals a certain grouping tendency such that the centers of gravity of these groups of points representing texts in different languages are shifted relative to each other. In more quantitative terms this can be read from the remaining seven panels where the group of points for each language is shown in separation with ellipses indicating the principal components of the corresponding points set. It is interesting to see that the points within a given set largely tend to be scattered in the $p-\beta$ plane along the isolines of constant expected value of the discrete Weibull distribution. This points to the existence of a characteristic scale for each language considered but that this scale varies among languages.

\subsection{Hurst exponents}

\begin{figure}
\centering
\captionsetup[subfloat]{margin=15pt}
\subfloat[All punctuation marks, books with one scaling regime; Rayuela: $H = 0.76$; Ogniem i mieczem: $H = 0.68$; Il piacere: $H = 0.59$; Sons and Lovers: $H = 0.66$]{\includegraphics[width=0.49\textwidth]{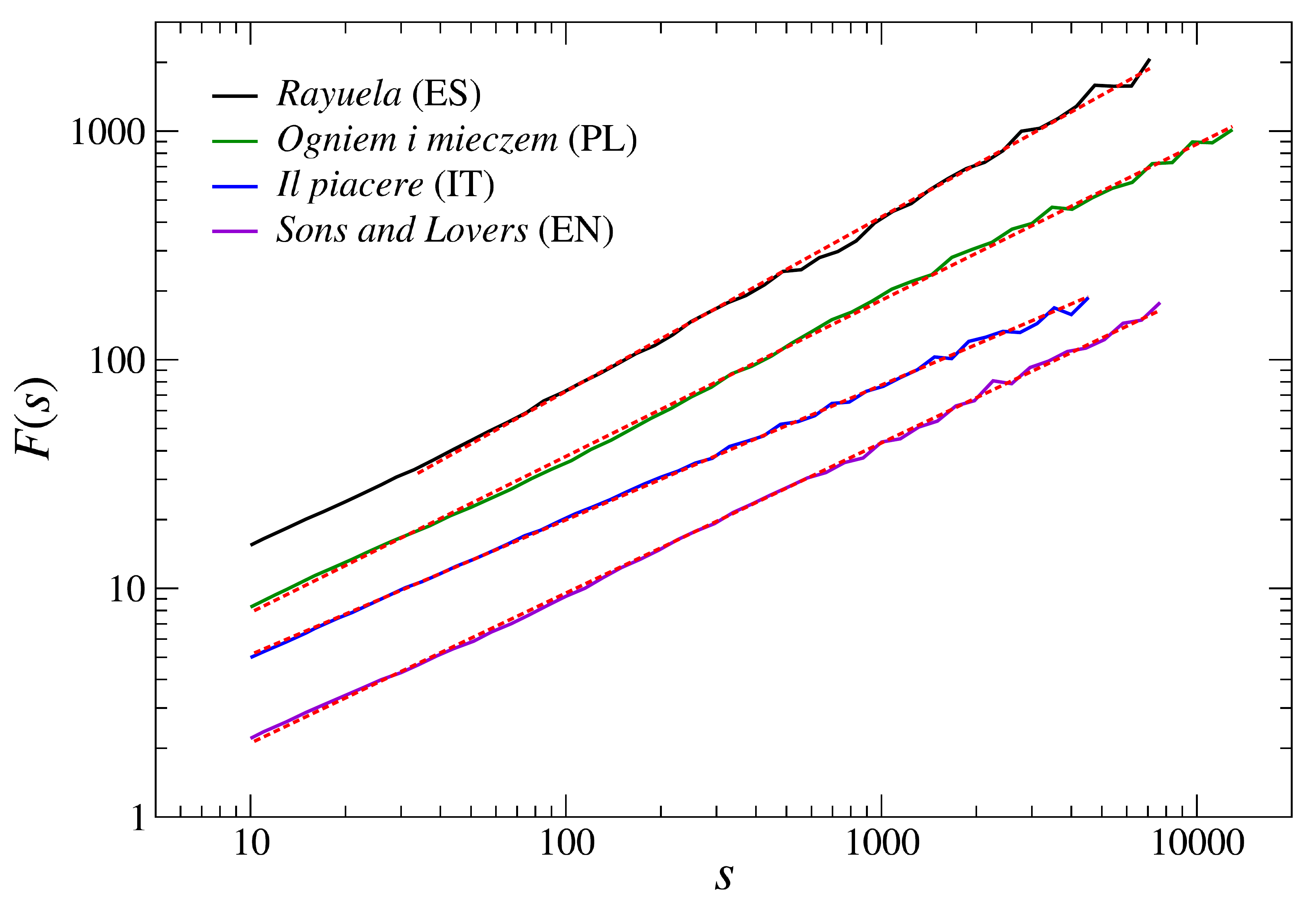}}
\subfloat[Sentence-ending marks, books with one scaling regime; Rayuela: $H = 0.72$; Ogniem i mieczem: $H = 0.74$; Il piacere: $H = 0.68$; Sons and Lovers: $H = 0.68$]{\includegraphics[width=0.49\textwidth]{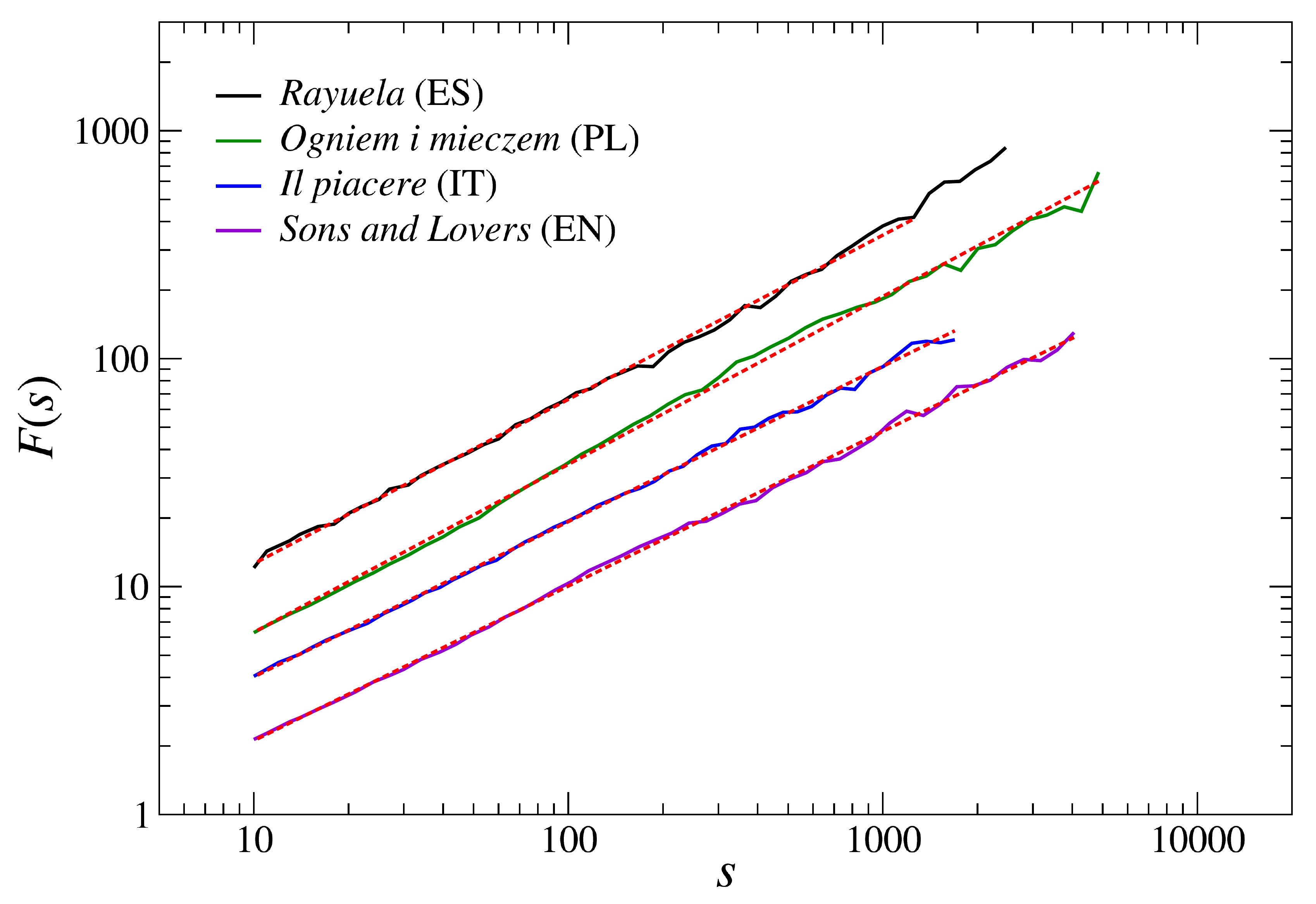}}
\vspace{2em}
\subfloat[All punctuation marks, books with two scaling regimes; The Goldfinch: $H^{I} = 0.57, H^{II} = 0.79$; Le Comte de Monte-Cristo: $H^{I} = 0.60, H^{II} = 0.72$; Quo Vadis: $H^{I} = 0.60, H^{II} = 0.68$; Bratya Karamazovy: $H^{I} = 0.59, H^{II} = 0.80$]{\includegraphics[width=0.49\textwidth]{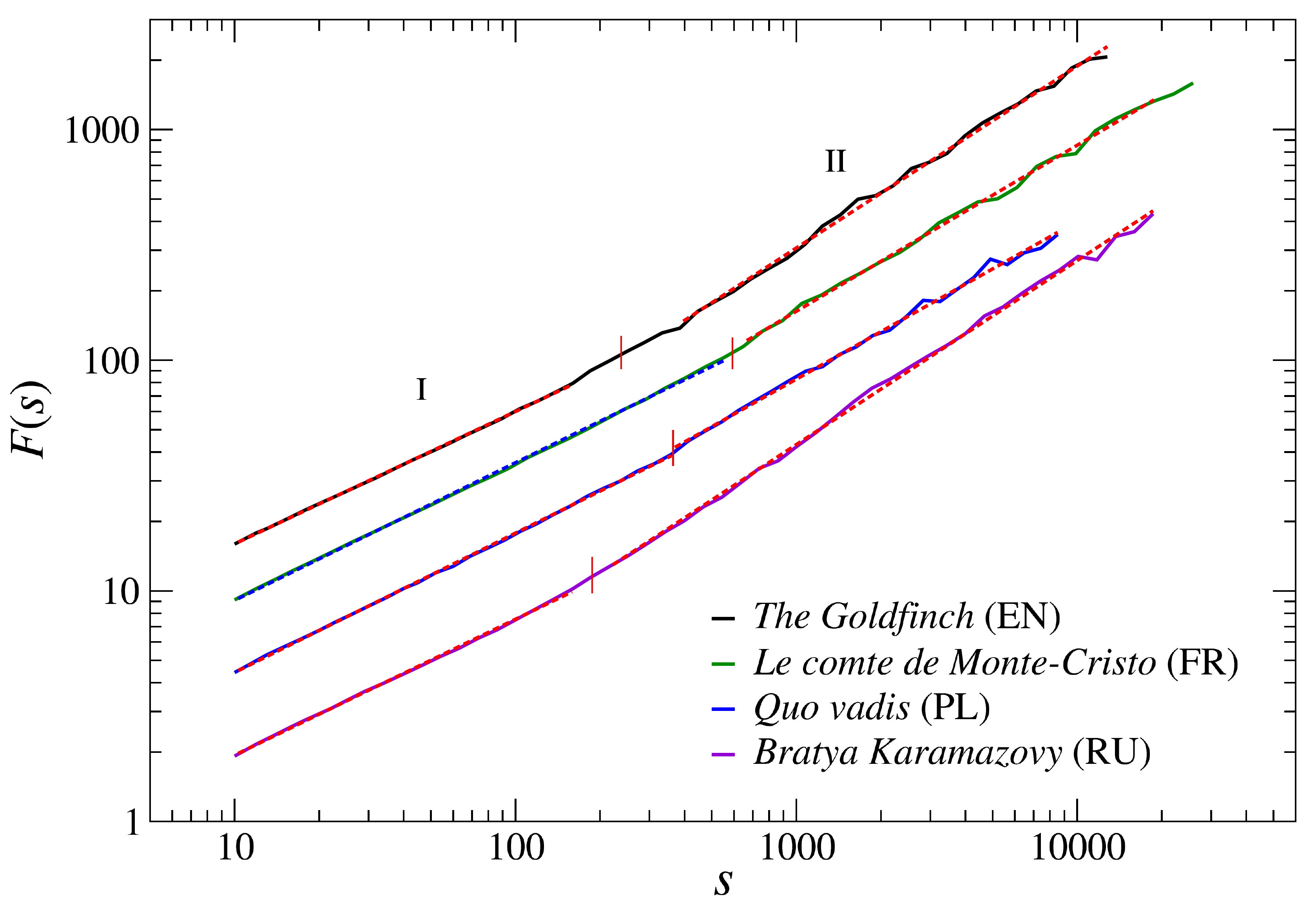}}
\subfloat[Sentence-ending marks, books with two scaling regimes; The Goldfinch: $H^{I} = 0.67, H^{II} = 0.76$; Le Comte de Monte-Cristo: $H^{I} = 0.75, H^{II} = 0.66$; Quo Vadis: $H^{I} = 0.72$ (no crossover); Bratya Karamazovy: $H^{I} = 0.67, H^{II} = 0.82$]{\includegraphics[width=0.49\textwidth]{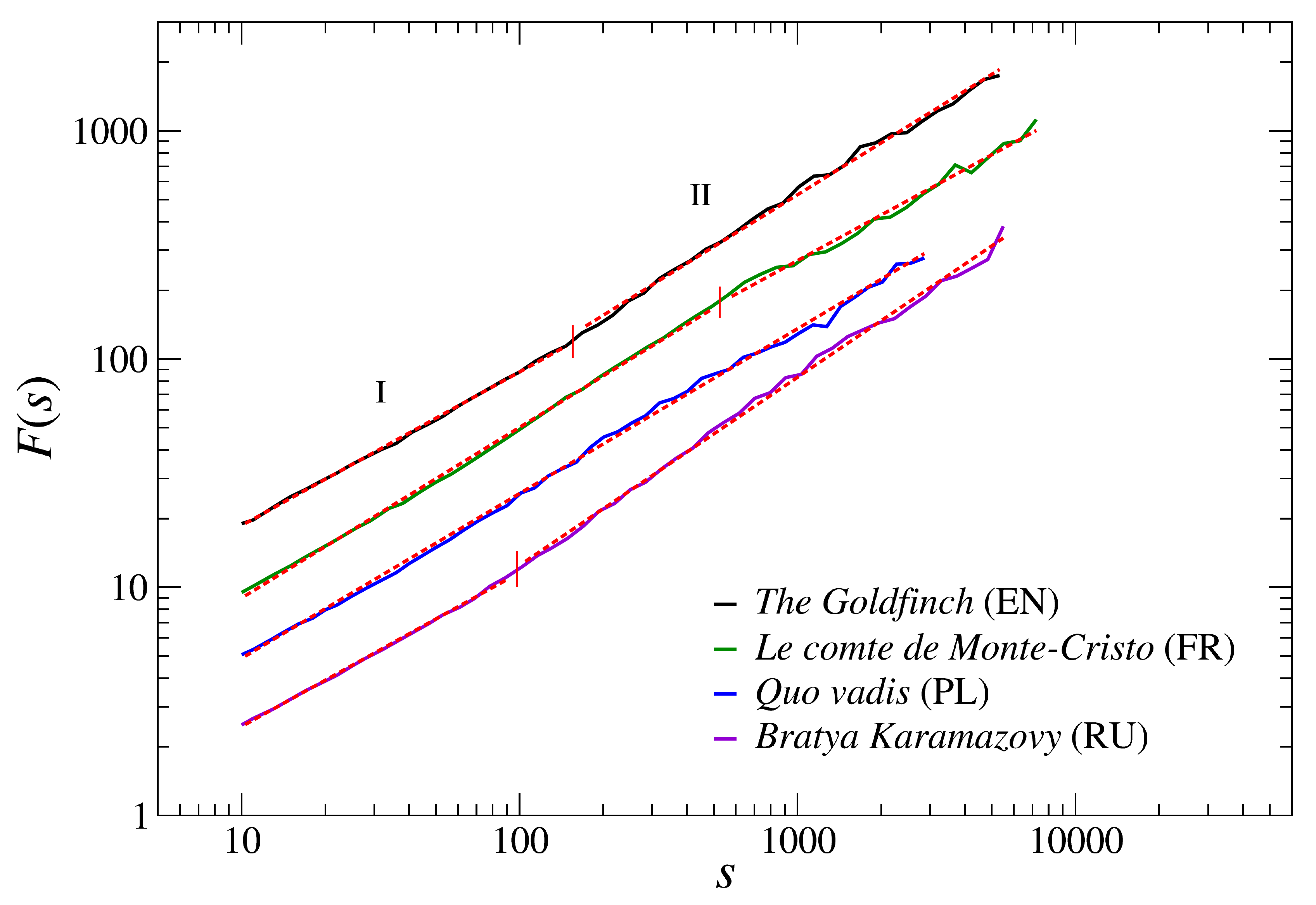}}
\caption{The DFA fluctuation functions $F(s)$ in the log-log scale for the times series of inter-punctuation intervals in the books indicated. Figures (a) and (c) show the results with all the punctuation marks included, and figures (b) and (d) show the results for sentence-ending marks only. The books presented in (a) and (b) are examples of texts whose punctuation has one $F(s)$ scaling regime with a single Hurst exponent $H$; in the books shown in (c) and (d) there is a crossover and two regimes with different Hurst exponents $H^{I}$ and $H^{II}$ are identified for the time series representing distances between consecutive punctuation marks. The values of $H$, $H^{I}$, and $H^{II}$ are given below each plot. The straight dashed lines indicate the best fits from which the Hurst exponents are determined.}
\label{fig::DFA.fluctuation.functions}
\end{figure}

Examples illustrating the typical behaviour of fluctuation functions $F(s)$ defined in \ref{sect::appendix-b} and computed for the series of inter-punctuation distances in the studied books are shown in Fig.~\ref{fig::DFA.fluctuation.functions}; two variants of time series are considered - one for all the punctuation included (all) and one for the sentence ending marks only (stop). Scaling appears to take place in essentially all the cases considered but in many of them at around the middle of scales there is a cross-over from one scaling region (I) to another (II), which thus results in two different Hurst exponents $H^{\rm I}$ and $H^{\rm II}$ with the latter being larger. This indicates that the effects of persistence are stronger at larger scales. In some of the books there is no such cross-over and one uniform scaling is developed with one Hurst exponent. It is also interesting to see that the quality of scaling is typically better when all the punctuation is taken into account (all) as compared to the sentence-ending marks (stop) only.

\begin{figure}
\centering
\includegraphics[width=0.75\textwidth]{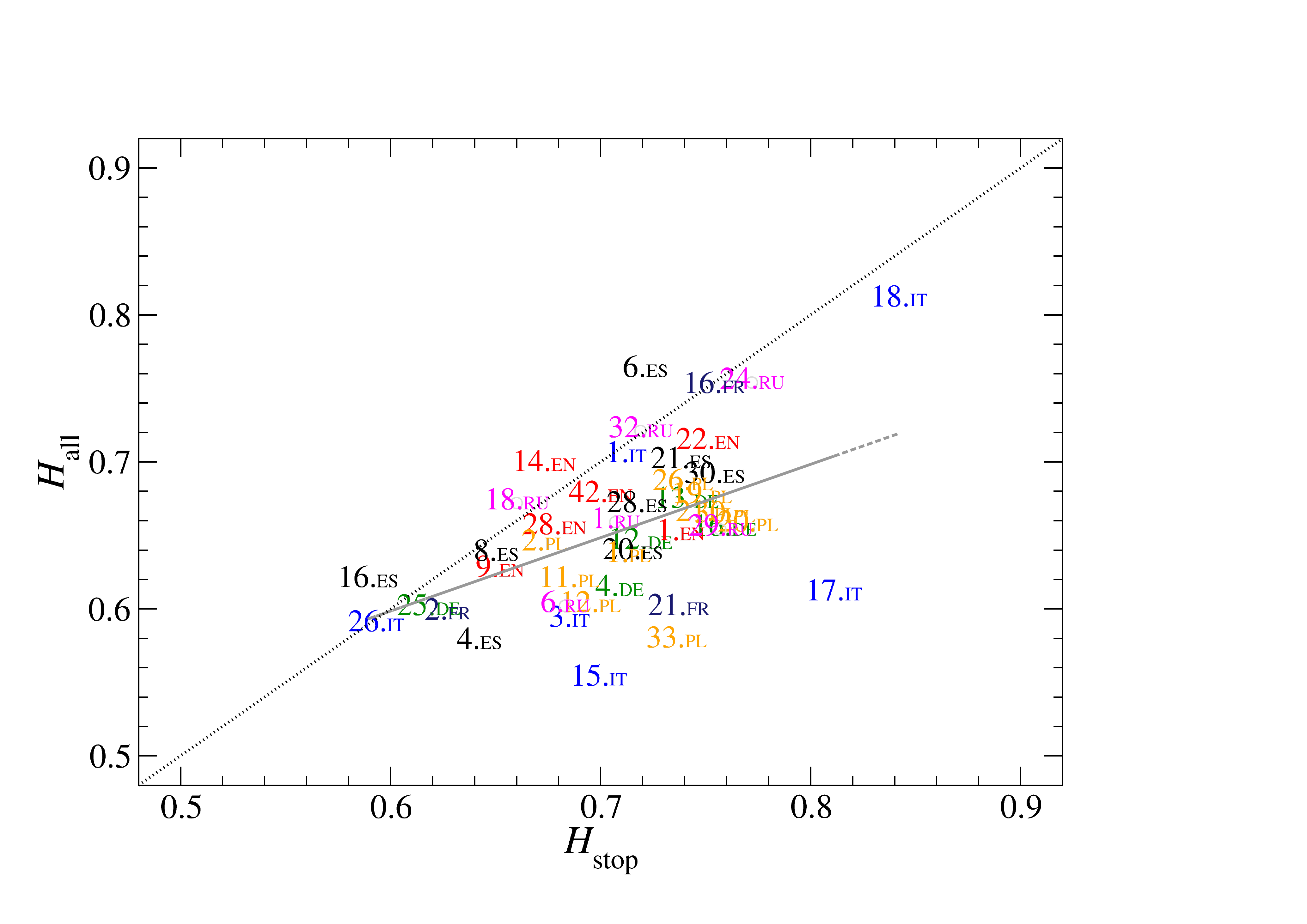}
\caption{The Hurst exponents calculated from the series of inter-punctuation intervals in the books indicated by the numbers from the list (see~\ref{sect::appendix-b}). Detrended fluctuation analysis~\cite{Peng1994} with the second order detrending polynomial is used for this purpose. Only these books are here included that develop a uniform unique  scaling of the corresponding fluctuation functions over the entire range of scales allowed by the length of a book.  Abscissa values reflect the Hurst exponents of the sentence-ending punctuation and ordinate for all the punctuation included. The straight line crossing the cloud of points represents the least-square fit.}
\label{fig::hurst.exponents}
\end{figure}

The scatter plot shown in Fig.~\ref{fig::hurst.exponents} collects pairs of Hurst exponents for a subset of texts that develop a unique scaling over the entire range of scales allowed by the length of text. For 44 such texts identified, the Hurst exponents are determined in two variants: one is for the series of sentence length (horizontal axis) and the other for the corresponding series of the consecutive intervals between all the punctuation (vertical axis). In all the cases the Hurst exponents appear larger than $0.5$ which indicates the existence of correlations of the persistent character. In the sentence-ending case, however, these exponents, on average, are significantly larger. This suggests that word sequences between consecutive punctuation marks are more constrained than sentences - in the sense that more freedom is allowed in the arrangemnt of sentences and in correlations between their lengths~\cite{Drozdz2016}. This seems to be consistent with the results regarding probability distributions.  As already discussed, the distances between consecutive punctuation marks can be described with one particular form of probability distribution (discrete Weibull distribution), while for sentence lengths such a description is significantly less accurate - therefore, sentence lengths can be said to have less constrained variability.

\begin{figure}
\centering
\includegraphics[width=0.65\textwidth]{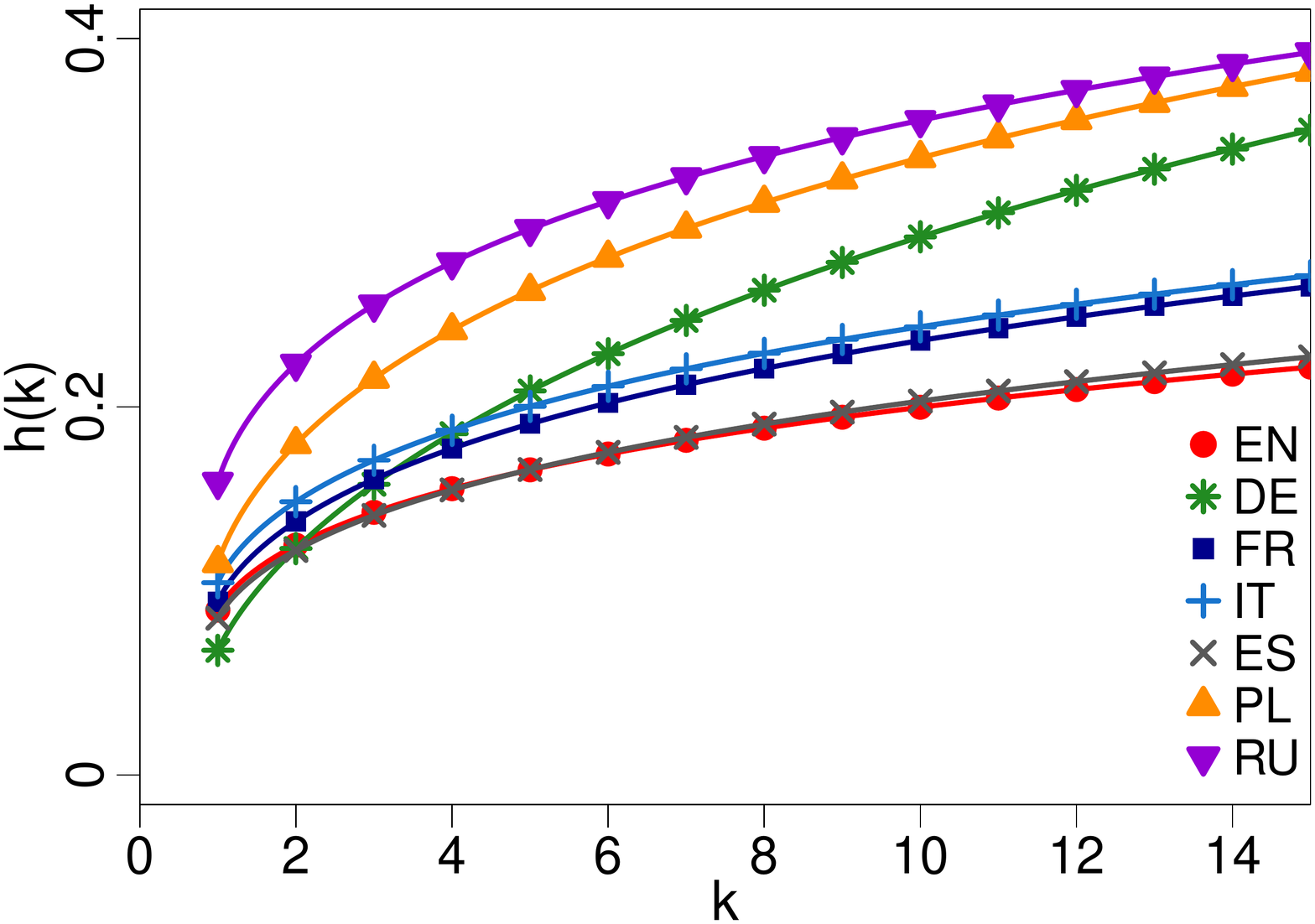}
\caption{The hazard functions $h(k)$ corresponding to the distributions of inter-punctuation intervals $k$ for all the books in each language separately. $h(k)$ are computed according to Eq.~(\ref{eq::hazard.Weibull}) using the average values $\langle p\rangle$ and $\langle\beta\rangle$ of the best fitting parameters $p$ and $\beta$ of discrete Weibull distributions.} 
\label{fig::hazard.functions}
\end{figure}

\subsection{Hazard rates}


\begin{figure}
\centering
\includegraphics[width=0.65\textwidth]{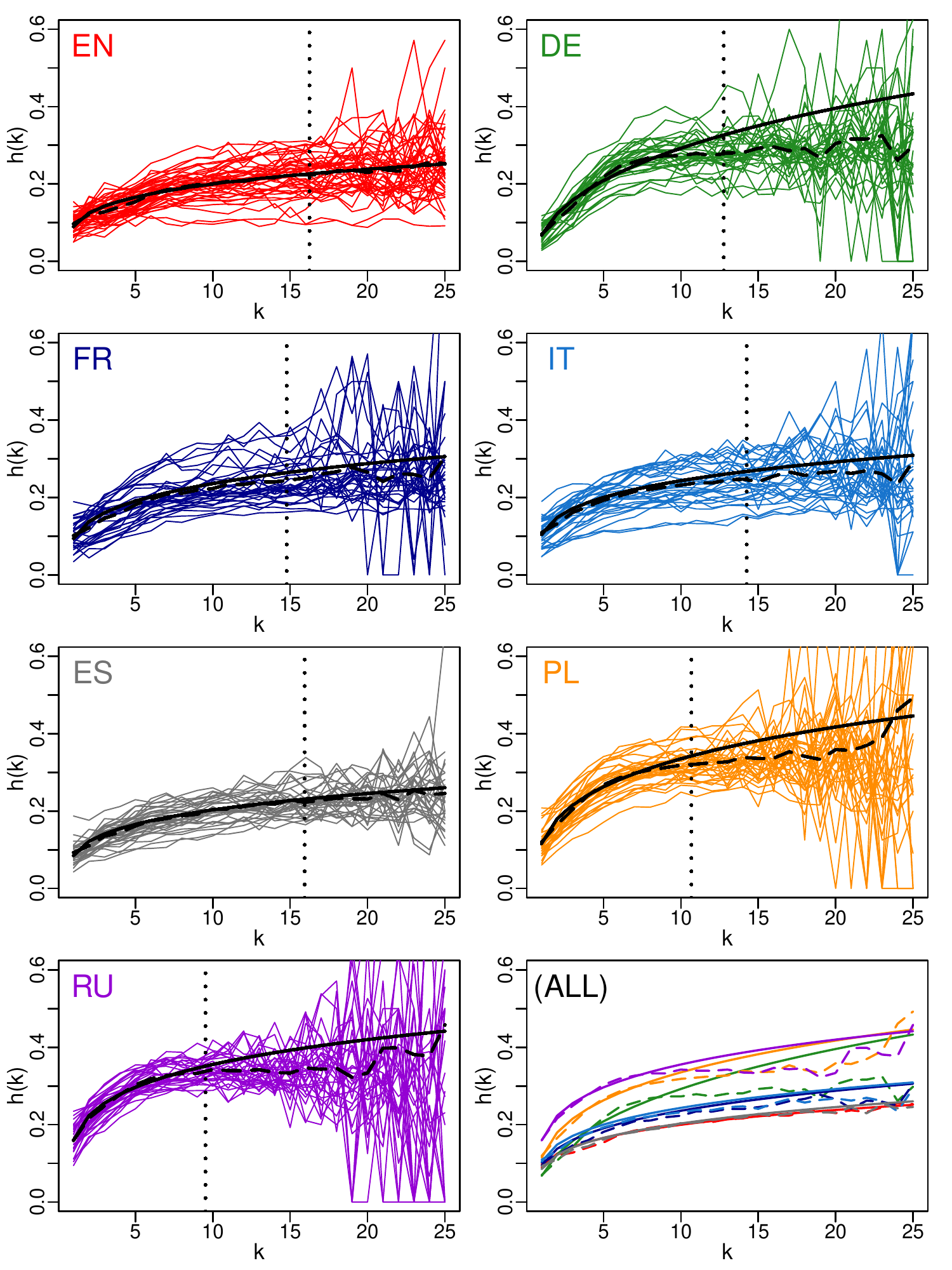}
\caption{The hazard functions $h(k)$ characterizing the distributions of the distances $k$ between punctuation marks in different languages, with an illustration of two different methods of averaging. In each of the first 7 plots (EN, DE, FR, IT, ES, PL, RU), each of the colored lines represents an empirical hazard function (a hazard function determined from the empirical distribution of punctuation recurrence distances) of a single book in the given language. The solid black line represents the hazard function of the discrete Weibull distribution with parameters $\langle p\rangle$ (average $p$ in the studied language) and $\langle\beta\rangle$ (average $\beta$ in the studied language); this is the same function as presented in Fig.~\ref{fig::hazard.functions}. The function represented by the dashed black line is obtained by averaging empirical hazard functions directly: the value of that function for a given $k$ is equal to the arithmetic mean of the values that the hazard functions describing individual books have for that $k$. The dotted vertical line marks the upper bound of the range of $k$ for which the results can be considered reliable. The bound is computed as the average value of the 95th percentile of the punctuation recurrence distance distribution (the 95th percentile of the distribution is determined for each of the books in a given language independently, and the obtained values are averaged). The large oscillations of the empirical hazard functions for large $k$ (outside the range of statistical ''reliability'') are a consequence of poor estimation of probability, which is natural in the tail of the empirical distribution. The last plot (ALL) presents the averaged hazard functions for each of the studied languages, computed by the two described methods; the dashed and the dotted lines correspond to particular methods in the same manner as in the remaining plots, the colors correspond to languages.}
\label{fig::average.hazard.functions}
\end{figure}

The distribution of punctuation patterns can succinctly and pictorially be comprised by the hazard functions of Eq.~(\ref{eq::hazard.Weibull}). In the present context they provide information how ''urgent'' it is to place a punctuation mark in order to pause an uninterrupted word sequence, depending on the length of that sequence. As it is seen from Eq.~(\ref{eq::hazard.Weibull}) the behaviour of $h(k)$ at the intermediate values of $k$ is determined by the interplay between the values of $p$ and $\beta$ parameters. It begins with $p$ at $k=1$ and its changes are governed by the value of $\beta$ parameter. In all the studied cases, $\beta$ is greater than 1, therefore $h(k)$ is an increasing function - which seems to be an expected result, as it indicates that the ''pressure'' to terminate a word sequence increases with the length of that sequence. Hazard functions $h(k)$ calculated from the average values of the parameters $p$ and $\beta$ for each corpus independently are displayed in Fig.~\ref{fig::hazard.functions} within the range of recurrence intervals $k$ between 1 and 15 (which corresponds to more than 80\% of all observed inter-punctuation intervals in each of the studied texts). This representation even more distinctly reveals the differences among the languages studied. In these average values by far the lowest hazard rate is revealed by the English texts and only slightly higher by the Spanish ones. This latter case is particularly striking in view of the fact that for the other two Romance languages~\cite{Posner1996}, French and Italian, $h(k)$ develops values significantly larger and, in addition, is very similar for both of them. It may seem less surprising that in the upper extreme there are the two Slavic languages, Polish and Russian. They are more inclined towards short word sequences between consecutive punctuation marks -- and, indeed, in the considered range of $k$ they are seen to systematically develop the values of $h(k)$ by about a factor of two larger than the ones in the lower extreme corresponding to English and Spanish. The averaged $h(k)$ for German, having the lowest $p$ and the highest $\beta$, behaves somewhat differently as a function of $k$. It starts from the lowest value but increases considerably faster and very quickly exceeds English, Spanish, French and Italian, and even tends to approach the two Slavic languages.

An issue that needs an extra test in this context is to what extent the shapes of $h(k)$ in Fig.~\ref{fig::hazard.functions} are forced by the functional form of Eq.~(\ref{eq::hazard.Weibull}). In order to address this point the hazard rates for all the texts are calculated also directly from the definition and averaged within each corpus. For English and Spanish there appears almost no difference (Fig.~\ref{fig::average.hazard.functions}) between the two ways of evaluating the hazard rate for as large values as $k \approx 25$ -- which points to a large stability and an internal consistency of this result. For the other languages this agreement is seen to terminate somewhat sooner but still acceptable within the statistically significant ranges of $k$.

\subsection{Text translation}

\begin{figure}
\centering
\includegraphics[width=1.0\linewidth]{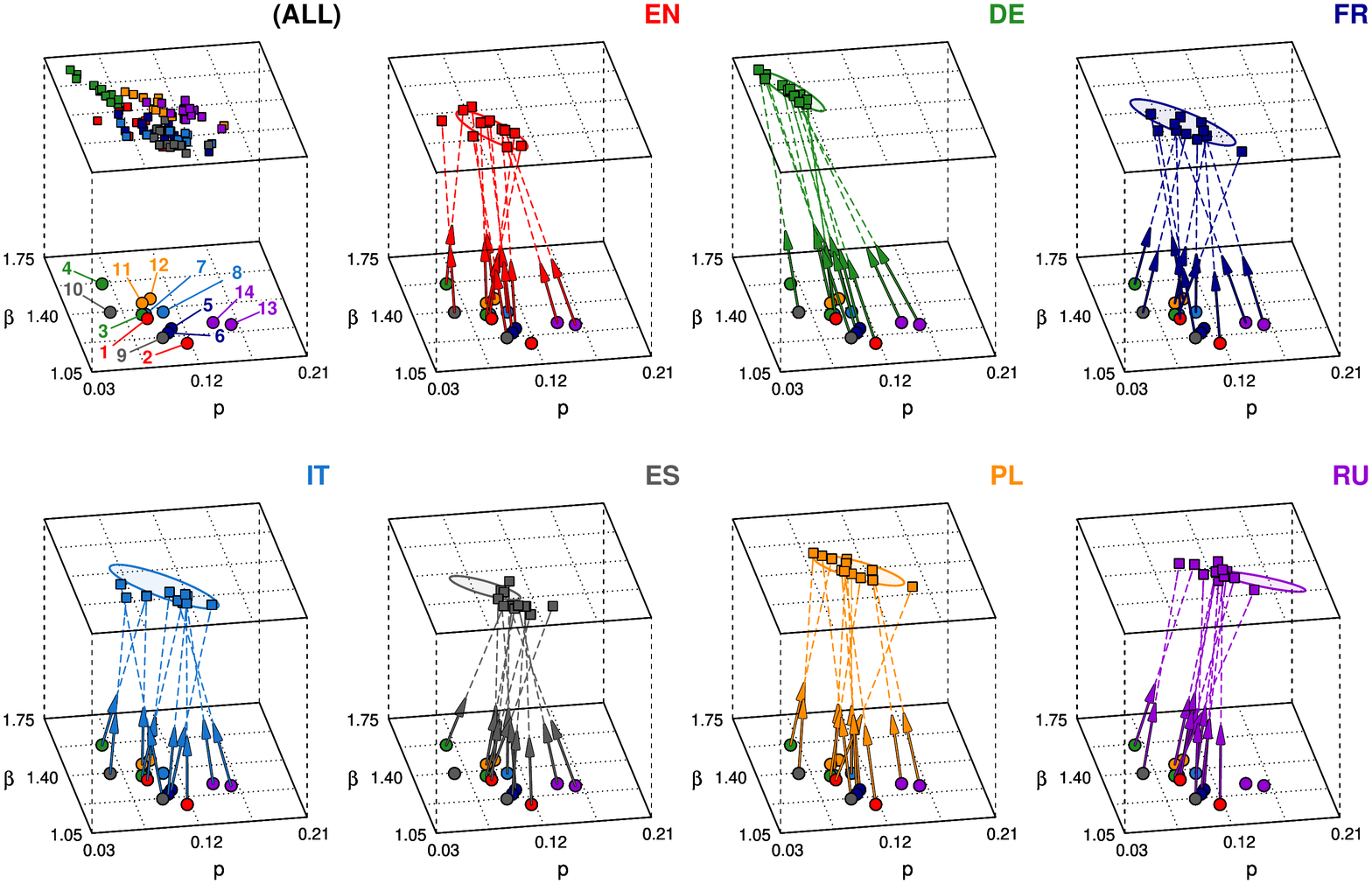}
\caption{The effect of text translation on the parameters $p,\beta$ of the discrete Weibull distribution describing the distances between consecutive punctuation marks. The figure labeled with ``(ALL)'' presents all the studied languages collectively, the remaining ones show the results for translations into individual target languages. The set of texts consists of 14 books and their translations: 2 books in each of the 7 studied languages are translated into 6 remaining languages; with two translations missing, this gives 96 texts in total. In each of the figures, the dots on the lower plane mark the values of $p$ and $\beta$ of the texts in their original languages, and the squares on the upper plane give $p$ and $\beta$ for the same texts translated into the target language (except the texts whose original language is the target language). The correspondence between the original and translated texts is marked by dashed lines and arrows (with the exception of the ``(ALL)'' plot). Colors of dots and squares represent individual languages in the same way as in Figs. \ref{fig::scatterplots.original} and \ref{fig::hazard.functions}. The values of $p$ and $\beta$ of the original 14 books are labeled in the ``(ALL)'' figure - the numbers correspond to the numbers given in~\ref{sect::appendix-c} (in the ''translated texts'' section). The ellipses on the upper plane are the same ellipses that are shown in Fig. \ref{fig::scatterplots.original}.}
\label{fig::3d.planes.scatterplots.translations}
\end{figure}

The system-specific characteristics of languages regarding their inter-punctuation patterns, as identified above, give raise to the question on how the related features transform under translation from one language to another. The present collection of great-quality original texts and their translations allows to reliably approach an answer. Fig.~\ref{fig::3d.planes.scatterplots.translations} illustrates how the $p$ and $\beta$ parameters of 7 pairs (14 in total) of texts representing all the seven languages studied transform under translation into the other six languages. Quite impressively, the translations are seen to navigate the resulting $p$ and $\beta$ parameters towards the ellipsoids of Fig.~\ref{fig::scatterplots.original} of the target languages. A more detailed information on the correspondence between the original and translated texts is given in~\ref{sect::appendix-c}. Even though this demonstration is limited in the statistical terms, it nevertheless already provides a strong indication for the robustness of the correlations observed.


\section{Summary and outlook}

Even though punctuation is almost commonly recognized important, often even crucial in properly conveying the intended meaning and facilitating structuring and transmission of information, so far its quantitative analyses are seldom in the scientific literature, especially on the quantitative level. The results shown here indicate direction and document that such a kind of study can be conducted and inspiring conclusions obtained, revealing both the universal as well as the system-specific aspects. Particularly intriguing is the result that among the major Western languages these are English and not much more Spanish that are the least demanding concerning the necessity to place the next punctuation mark. Still they both are definitely not less precise in formulating utterances than the other languages. Such observations may be taken as an indication that these two languages are better oriented towards optimizing communicative success and expressiveness by, at the same time, reducing the physical and cognitive effort involved. The fact that out of the languages here considered this appears to apply just to the two most global world languages~\cite{Crystal2003,Ronen2014} may not be accidental but points to some deeper foundation. In this context it is advisable to keep in mind that, whatever the mechanism of language is, its functioning is embedded in biological environment and is thus exposed to obey the related biological constraints and preferences~\cite{Evans2009,Lupyan2016}. Concepts like generative grammar attempt to explain competence but not performance; some languages may facilitate performance. During their development languages have been gradually reshaping and adopting themselves under influence of other languages and dialects they have encountered. Especially English and Spanish experienced many of such encounters. Whether this made them as they are or they became global because they happen to possess some spontaneously acquired favorable intrinsic syntactic organization emerges an exciting issue for further interdisciplinary explorations.

\bibliographystyle{elsarticle-num}

\newpage

\appendix

\section{Weibull plots $(x,y)$ and rescaled Weibull plots $(\widetilde{x}, \widetilde{y})$}
\label{sect::appendix-a}

A practical way of assessing and visualizing how well a given data set fits to a Weibull distribution is constructing the so-called Weibull plot. For this purpose Eq.~(\ref{eq::DW.CCDF}) is rewritten in the form:
\begin{equation}
\log \left( -\log \left( 1-F(k) \right) \right) = \beta \log k + \log \left( -\log \left( 1-p \right) \right).
\label{eq::Weibull.linearization}
\end{equation}
Then, for the data originating from the discrete Weibull distribution with parameters $(p,\beta)$, a plot of the empirical cumulative distribution function $F_{emp}(k)$ in coordinates $(x,y)$, where
\begin{equation} 
x=\log k, \qquad y=\log\left( -\log \left( 1-F_{emp}(k) \right) \right),
\label{eq::Weibull.plot}
\end{equation}
results in  a straight line with slope $\beta$ and intercept $\log\left( -\log \left( 1-p \right) \right)$. 

To obtain a rescaled plot, which fits in the square $[0,1]\!\times\![0,1]$, and has the reference line with slope~1 and intercept~0, one applies the following transformation. Let $y=a+bx$ denote the equation describing the line representing the assumed Weibull distribution in a considered Weibull plot. The boundaries of the rectangle enclosing the $y(x)$ plot, denoted by $x_{\rm plot.min}$, $x_{\rm plot.max}$, $y_{\rm plot.min}$, $y_{\rm plot.max}$, are defined as:
\begin{eqnarray}
\nonumber
x_{\rm plot.min} = \min \left\{ x_{\rm min}, \, \frac{y_{\rm min}-a}{b} \right\}, \qquad x_{\rm plot.max} = \max \left\{ x_{\rm max}, \, \frac{y_{\rm max}-a}{b} \right\}, \\
y_{\rm plot.min} = \min \left\{ y_{\rm min}, \, a + b\,x_{\rm min} \right\}, \qquad y_{\rm plot.max} = \max \left\{ y_{\rm max}, \, a + b\,x_{\rm max} \right\}.
\label{eq::rescaled.Weibull.plot.transformation.1}
\end{eqnarray}
The transformation from the Weibull plot $(x,y)$ to the rescaled Weibull plot $(\widetilde{x},\widetilde{y})$ is then  given by:
\begin{equation}
\widetilde{x}=\frac{x-x_{\rm plot.min}}{x_{\rm plot.max}-x_{\rm plot.min}}, \qquad \widetilde{y}=\frac{y-y_{\rm plot.min}}{y_{\rm plot.max}-y_{\rm plot.min}}.
\label{eq::rescaled.Weibull.plot.transformation.2}
\end{equation}

\newenvironment{packed_enum}{
\begin{enumerate}
  \setlength{\itemsep}{1pt}
  \setlength{\parskip}{0pt}
  \setlength{\parsep}{0pt}
}{\end{enumerate}}

\vspace{4em}

\section{Detrended Fluctuation Analysis (DFA)}
\label{sect::appendix-b}

Computing the Hurst exponent of a real-valued time series $x(1), x(2), x(3),..., x(N)$ with the use of DFA can be presented in the form of an algorithm consisting of the following steps.

\begin{enumerate}
\item Determine the \textit{profile} of the series, that is, the cumulative series: 
\begin{equation*}
y(k) = \sum_{i=1}^{k} x(i).
\end{equation*}

\item Divide the series into $N_s = \lfloor N/s \rfloor$ non-overlapping segments $I_{\nu}$ of equal length $s$ (the notation $\lfloor \boldsymbol{\cdot} \rfloor$ represents the floor function, and $I_{\nu}$ denotes the set of time series' indices belonging to the $\nu$-th segment; a segment of length $s$ starting at some index $k$ consists of indices $k, k + 1, k + 2, ..., k + s - 1$). To avoid disregarding any piece of the data, make two partitions - one starting from $k = 1$ and one starting from the opposite end of the series; this gives $2N_s$ (overlapping) segments in total.

\item \textit{Detrend} the cumulative series - compute a local polynomial trend $p_\nu$ for each segment $I_\nu$, using the least squares method and construct the detrended cumulative series $y(k) - p_\nu(k)$. The order of the chosen polynomial influences the shape of the trend that can be removed from the data; common choices are linear, quadratic, and cubic polynomials.

\item Compute the quantity $F^2(\nu,s)$, called variance, for each segment $I_\nu$:
\begin{equation*}
F^2(\nu,s) = \frac{1}{s}\sum_{k \in I_\nu} \!\! \left( y(k) - p_\nu(k) \right)^2.
\end{equation*}

\item Compute a single value of the \textit{fluctuation function} $F(s)$ (which characterizes the fluctuations at a given scale) for a given $s$ as the square root of the mean of the variances $F^2(\nu,s)$:
\begin{equation*}
F(s) = \left( \dfrac{1}{2N_s} \sum\limits_{\nu=1}^{2N_s} F^2(\nu,s)\right)^{1/2}.
\end{equation*}

\item Repeat the steps from 2 to 5 for different segment lengths $s$ chosen from some range $[s_{\min}; s_{\max}]$; the choice depends on the studied data, but it is often suggested that $s_{\min}$ should be not less than 10 and $s_{\max}$ not greater than $N/5$. With that procedure, the set of values of $F(s)$ for different $s$ is obtained. 

\item Investigate the scaling behavior of $F(s)$; if within the studied range of $s$ the fluctuation function is described by a power law of the form:
\begin{equation*}
F(s) \approx Cs^{H},
\end{equation*}
where $C$ is a constant, then $H$ is the Hurst exponent characterizing the studied time series.
\end{enumerate}

\vspace{4em}

\section{Books used in the study}
\label{sect::appendix-c}

All the texts used in the analysis are given below; the list contains titles, authors' names, and the parameters $p$ and $\beta$ of the discrete Weibull distribution fitted to the intervals between merged punctuation marks. The texts used to study the effects of translation are listed below the main data set; these texts are assigned numbers, corresponding to the numbers labeling points in Figs.~\ref{fig::hurst.exponents} and~\ref{fig::3d.planes.scatterplots.translations}.

\vspace{0.3cm}

\noindent
{\bf The main data set}

\vspace{0.5\baselineskip}

\noindent\underline{English}

\begin{packed_enum}
\item \textit{Little Women} - L. M. Alcott; $p\!=\!0.07$, $\beta\!=\!1.37$
\item \textit{Pride and Prejudice} - J. Austen; $p\!=\!0.06$, $\beta\!=\!1.40$
\item \textit{Sense and Sensibility} - J. Austen; $p\!=\!0.06$, $\beta\!=\!1.43$
\item \textit{Jane Eyre: An Autobiography} - C. Brontë; $p\!=\!0.10$, $\beta\!=\!1.30$
\item \textit{The Thirty-Nine Steps} - J. Buchan; $p\!=\!0.03$, $\beta\!=\!1.58$
\item \textit{The Adventures of Tom Sawyer} - S. L. Clemens; $p\!=\!0.13$, $\beta\!=\!1.11$
\item \textit{The Prince and the Pauper} - S. L. Clemens; $p\!=\!0.09$, $\beta\!=\!1.27$
\item \textit{The Rotters' Club} - J. Coe; $p\!=\!0.11$, $\beta\!=\!1.16$
\item \textit{Heart of Darkness} - J. Conrad; $p\!=\!0.10$, $\beta\!=\!1.23$
\item \textit{The Last of the Mohicans} - J. F. Cooper; $p\!=\!0.07$, $\beta\!=\!1.33$
\item \textit{Robinson Crusoe} - D. Defoe; $p\!=\!0.05$, $\beta\!=\!1.55$
\item \textit{David Copperfield} - C. Dickens; $p\!=\!0.09$, $\beta\!=\!1.34$
\item \textit{Olivier Twist} - C. Dickens; $p\!=\!0.12$, $\beta\!=\!1.25$
\item \textit{The Pickwick Papers} - C. Dickens; $p\!=\!0.14$, $\beta\!=\!1.17$
\item \textit{Alice's Adventures in Wonderland} - C. L. Dodgson; $p\!=\!0.09$, $\beta\!=\!1.38$
\item \textit{The Adventures of Sherlock Holmes} - A. C. Doyle; $p\!=\!0.10$, $\beta\!=\!1.15$
\item \textit{Middlemarch} - M. A. Evans; $p\!=\!0.07$, $\beta\!=\!1.25$
\item \textit{Absalom, Absalom!} - W. Faulkner; $p\!=\!0.11$, $\beta\!=\!1.00$
\item \textit{The History of Tom Jones, a Foundling} - H. Fielding; $p\!=\!0.09$, $\beta\!=\!1.24$
\item \textit{The Great Gatsby} - F. S. K. Fitzgerald; $p\!=\!0.09$, $\beta\!=\!1.23$
\item \textit{A Room with a View} - E. M. Forster; $p\!=\!0.10$, $\beta\!=\!1.34$
\item \textit{Tess of the d’Urbervilles} - T. Hardy; $p\!=\!0.09$, $\beta\!=\!1.23$
\item \textit{The Scarlet Letter} - N. Hawthorne; $p\!=\!0.09$, $\beta\!=\!1.28$
\item \textit{The Old Man and the Sea} - E. Hemingway; $p\!=\!0.07$, $\beta\!=\!1.13$
\item \textit{Brave New World} - A. Huxley; $p\!=\!0.12$, $\beta\!=\!1.21$
\item \textit{The Portrait of a Lady} - H. James; $p\!=\!0.07$, $\beta\!=\!1.34$
\item \textit{Kim} - J. R. Kipling; $p\!=\!0.09$, $\beta\!=\!1.30$
\item \textit{Sons and Lovers} - D. H. Lawrence; $p\!=\!0.09$, $\beta\!=\!1.40$
\item \textit{The Lion, the Witch and the Wardrobe} - C. S. Lewis; $p\!=\!0.12$, $\beta\!=\!1.07$
\item \textit{Moby-Dick; or, The Whale} - H. Melville; $p\!=\!0.11$, $\beta\!=\!1.22$
\item \textit{Gone with the Wind} - M. Mitchell; $p\!=\!0.10$, $\beta\!=\!1.12$
\item \textit{Anne of Green Gables} - L. M. Montgomery; $p\!=\!0.10$, $\beta\!=\!1.13$
\item \textit{Animal Farm} - G. Orwell; $p\!=\!0.06$, $\beta\!=\!1.33$
\item \textit{Pointed Roofs} - D. Richardson; $p\!=\!0.12$, $\beta\!=\!1.11$
\item \textit{The Catcher in the Rye} - J. D. Salinger; $p\!=\!0.11$, $\beta\!=\!1.22$
\item \textit{Ivanhoe} - W. Scott; $p\!=\!0.09$, $\beta\!=\!1.29$
\item \textit{Old Mortality} - W. Scott; $p\!=\!0.09$, $\beta\!=\!1.24$
\item \textit{Frankenstein; or, The Modern Prometheus} - M. Shelley; $p\!=\!0.06$, $\beta\!=\!1.33$
\item \textit{Tristram Shandy} - L. Sterne; $p\!=\!0.09$, $\beta\!=\!1.35$
\item \textit{Treasure Island} - R. L. Stevenson; $p\!=\!0.12$, $\beta\!=\!1.18$
\item \textit{Gulliver's Travels} - J. Swift; $p\!=\!0.04$, $\beta\!=\!1.61$
\item \textit{Vanity Fair} - W. M. Thackeray; $p\!=\!0.08$, $\beta\!=\!1.29$
\item \textit{The Time Machine} - H. G. Wells; $p\!=\!0.06$, $\beta\!=\!1.43$
\item \textit{The Waves} - A. V. Woolf; $p\!=\!0.10$, $\beta\!=\!1.25$
\end{packed_enum}

\vspace{0.5\baselineskip}
\noindent\underline{German}

\begin{packed_enum}
\item \textit{Geschichte vom braven Kasperl und dem schönen Annerl} - C. Brentano; $p\!=\!0.06$, $\beta\!=\!1.55$
\item \textit{Wir Kinder vom Bahnhof Zoo} - C. Felscherinow; $p\!=\!0.03$, $\beta\!=\!1.84$
\item \textit{Der Stechlin} - T. Fontane; $p\!=\!0.08$, $\beta\!=\!1.45$
\item \textit{Effi Briest} - T. Fontane; $p\!=\!0.08$, $\beta\!=\!1.46$
\item \textit{Frau Jenny Treibel} - T. Fontane; $p\!=\!0.09$, $\beta\!=\!1.40$
\item \textit{Die Blechtrommel} - G. Grass; $p\!=\!0.07$, $\beta\!=\!1.50$
\item \textit{Klein Zaches genannt Zinnober} - E. T. A. Hoffmann; $p\!=\!0.08$, $\beta\!=\!1.40$
\item \textit{Lebens-Ansichten des Katers Murr} - E. T. A. Hoffmann; $p\!=\!0.07$, $\beta\!=\!1.52$
\item \textit{Das Schloss} - F. Kafka; $p\!=\!0.06$, $\beta\!=\!1.59$
\item \textit{Der Process} - F. Kafka; $p\!=\!0.05$, $\beta\!=\!1.64$
\item \textit{Emil und die Detektive} - E. Kästner; $p\!=\!0.08$, $\beta\!=\!1.51$
\item \textit{Der grüne Heinrich} - G. Keller; $p\!=\!0.03$, $\beta\!=\!1.67$
\item \textit{Die Leute von Seldwyla} - G. Keller; $p\!=\!0.04$, $\beta\!=\!1.62$
\item \textit{Der Untertan} - H. Mann; $p\!=\!0.06$, $\beta\!=\!1.65$
\item \textit{Professor Unrat oder Das Ende eines Tyrannen} - H. Mann; $p\!=\!0.06$, $\beta\!=\!1.60$
\item \textit{Buddenbrooks} - T. Mann; $p\!=\!0.09$, $\beta\!=\!1.37$
\item \textit{Der Zauberberg} - T. Mann; $p\!=\!0.07$, $\beta\!=\!1.47$
\item \textit{Doktor Faustus} - T. Mann; $p\!=\!0.07$, $\beta\!=\!1.47$
\item \textit{Winnetou} - K. May; $p\!=\!0.06$, $\beta\!=\!1.58$
\item \textit{Der Mann ohne Eigenschaften} - R. Musil; $p\!=\!0.04$, $\beta\!=\!1.67$
\item \textit{Also sprach Zarathustra} - F. Nietzsche; $p\!=\!0.07$, $\beta\!=\!1.56$
\item \textit{Die Chronik der Sperlingsgasse} - W. Raabe; $p\!=\!0.10$, $\beta\!=\!1.33$
\item \textit{Eulenpfingsten} - W. Raabe; $p\!=\!0.09$, $\beta\!=\!1.27$
\item \textit{Im Siegeskranze} - W. Raabe; $p\!=\!0.04$, $\beta\!=\!1.54$
\item \textit{Pfisters Mühle} - W. Raabe; $p\!=\!0.10$, $\beta\!=\!1.21$
\item \textit{Hiob} - J. Roth; $p\!=\!0.06$, $\beta\!=\!1.62$
\item \textit{Radetzkymarsch} - J. Roth; $p\!=\!0.07$, $\beta\!=\!1.54$
\item \textit{Der Schimmelreiter} - T. Storm; $p\!=\!0.08$, $\beta\!=\!1.38$
\item \textit{Immensee} - T. Storm; $p\!=\!0.06$, $\beta\!=\!1.56$
\item \textit{Peter Schlemihls wundersame Geschichte} - A. von Chamisso; $p\!=\!0.05$, $\beta\!=\!1.65$
\item \textit{Die Judenbuche – Ein Sittengemälde aus dem gebirgichten Westfalen} - A. von Droste-Hülshoff; $p\!=\!0.08$, $\beta\!=\!1.42$
\item \textit{Die Leiden des jungen Werthers} - J. W. von Goethe; $p\!=\!0.07$, $\beta\!=\!1.57$
\item \textit{Wilhelm Meisters Lehrjahre} - J. W. von Goethe; $p\!=\!0.05$, $\beta\!=\!1.63$
\item \textit{Michael Kohlhaas} - H. von Kleist; $p\!=\!0.10$, $\beta\!=\!1.52$
\end{packed_enum}

\vspace{0.5\baselineskip}
\noindent\underline{French}

\begin{packed_enum}
\item \textit{Candide} - F. M. Arouet; $p\!=\!0.09$, $\beta\!=\!1.31$
\item \textit{Le Rouge et le Noir, Chronique du XIXe siècle} - H. Beyle; $p\!=\!0.06$, $\beta\!=\!1.40$
\item \textit{La Chute} - A. Camus; $p\!=\!0.13$, $\beta\!=\!1.20$
\item \textit{La Peste} - A. Camus; $p\!=\!0.10$, $\beta\!=\!1.23$
\item \textit{Adolphe} - B. Constant; $p\!=\!0.05$, $\beta\!=\!1.51$
\item \textit{Atala, ou Les Amours de deux sauvages dans le désert} - F. R. de Chateaubriand; $p\!=\!0.04$, $\beta\!=\!1.65$
\item \textit{Germinie Lacerteux} - J. \& E. de Goncourt; $p\!=\!0.10$, $\beta\!=\!1.25$
\item \textit{La Princesse de Clèves} - M. M. de La Fayette; $p\!=\!0.05$, $\beta\!=\!1.40$
\item \textit{Les Liaisons dangereuses} - P. C. de Laclos; $p\!=\!0.06$, $\beta\!=\!1.46$
\item \textit{Bel-Ami} - G. de Maupassant; $p\!=\!0.13$, $\beta\!=\!1.20$
\item \textit{Le Petit Prince} - A. de Saint-Exupéry; $p\!=\!0.08$, $\beta\!=\!1.47$
\item \textit{Vol de nuit} - A. de Saint-Exupéry; $p\!=\!0.10$, $\beta\!=\!1.44$
\item \textit{Artamène ou le Grand Cyrus} - M. de Scudéry; $p\!=\!0.03$, $\beta\!=\!1.78$
\item \textit{La Reine Margot} - A. Dumas; $p\!=\!0.18$, $\beta\!=\!1.02$
\item \textit{Le Comte de Monte-Cristo} - A. Dumas; $p\!=\!0.14$, $\beta\!=\!1.10$
\item \textit{Les Trois Mousquetaires} - A. Dumas; $p\!=\!0.14$, $\beta\!=\!1.11$
\item \textit{Jacques} - A. A. L. Dupin; $p\!=\!0.06$, $\beta\!=\!1.40$
\item \textit{La Petite Fadette} - A. A. L. Dupin; $p\!=\!0.06$, $\beta\!=\!1.35$
\item \textit{Valentine} - A. A. L. Dupin; $p\!=\!0.08$, $\beta\!=\!1.27$
\item \textit{L’Éducation sentimentale} - G. Flaubert; $p\!=\!0.12$, $\beta\!=\!1.34$
\item \textit{Madame Bovary} - G. Flaubert; $p\!=\!0.12$, $\beta\!=\!1.28$
\item \textit{Le Grand Meaulnes} - H. A. Fournier; $p\!=\!0.09$, $\beta\!=\!1.30$
\item \textit{Mademoiselle de Maupin} - T. Gautier; $p\!=\!0.06$, $\beta\!=\!1.36$
\item \textit{Les Misérables} - V. Hugo; $p\!=\!0.10$, $\beta\!=\!1.25$
\item \textit{Notre-Dame de Paris} - V. Hugo; $p\!=\!0.12$, $\beta\!=\!1.18$
\item \textit{Le Fantôme de l'Opéra} - G. Leroux; $p\!=\!0.14$, $\beta\!=\!1.08$
\item \textit{Histoire du chevalier Des Grieux et de Manon Lescaut} - A. Prévost; $p\!=\!0.06$, $\beta\!=\!1.40$
\item \textit{À la recherche du temps perdu} - M. Proust; $p\!=\!0.08$, $\beta\!=\!1.24$
\item \textit{Le Tour du monde en quatre-vingts jours} - J. Verne; $p\!=\!0.11$, $\beta\!=\!1.27$
\item \textit{Vingt Mille Lieues sous les mers} - J. Verne; $p\!=\!0.10$, $\beta\!=\!1.26$
\item \textit{Lourdes} - É. Zola; $p\!=\!0.10$, $\beta\!=\!1.42$
\item \textit{Nana} - É. Zola; $p\!=\!0.13$, $\beta\!=\!1.34$
\end{packed_enum}

\vspace{0.5\baselineskip}
\noindent\underline{Italian}

\begin{packed_enum}
\item \textit{Il marchese di Roccaverdina} - L. Capuana; $p\!=\!0.14$, $\beta\!=\!1.14$
\item \textit{Le avventure di Pinocchio} - C. Collodi; $p\!=\!0.09$, $\beta\!=\!1.37$
\item \textit{Il piacere} - G. D'Annunzio; $p\!=\!0.13$, $\beta\!=\!1.16$
\item \textit{Ettore Fieramosca} - M. d'Azeglio; $p\!=\!0.06$, $\beta\!=\!1.42$
\item \textit{Niccolò dei Lapi} - M. d'Azeglio; $p\!=\!0.08$, $\beta\!=\!1.34$
\item \textit{Cuore} - E. De Amicis; $p\!=\!0.11$, $\beta\!=\!1.33$
\item \textit{Il romanzo di un maestro} - E. De Amicis; $p\!=\!0.08$, $\beta\!=\!1.37$
\item \textit{I Viceré} - F. De Roberto; $p\!=\!0.10$, $\beta\!=\!1.31$
\item \textit{L'Illusione} - F. De Roberto; $p\!=\!0.12$, $\beta\!=\!1.34$
\item \textit{Canne al vento} - G. Deledda; $p\!=\!0.14$, $\beta\!=\!1.13$
\item \textit{Cenere} - G. Deledda; $p\!=\!0.14$, $\beta\!=\!1.20$
\item \textit{Il nome della rosa} - U. Eco; $p\!=\!0.08$, $\beta\!=\!1.28$
\item \textit{Malombra} - A. Fogazzaro; $p\!=\!0.14$, $\beta\!=\!1.13$
\item \textit{Piccolo mondo antico} - A. Fogazzaro; $p\!=\!0.13$, $\beta\!=\!1.12$
\item \textit{Dell'arte della guerra} - N. Machiavelli; $p\!=\!0.06$, $\beta\!=\!1.40$
\item \textit{Il principe} - N. Machiavelli; $p\!=\!0.04$, $\beta\!=\!1.70$
\item \textit{I promessi sposi} - A. Manzoni; $p\!=\!0.11$, $\beta\!=\!1.37$
\item \textit{Le confessioni d'un italiano} - I. Nievo; $p\!=\!0.05$, $\beta\!=\!1.48$
\item \textit{Le mie prigioni} - S. Pellico; $p\!=\!0.07$, $\beta\!=\!1.53$
\item \textit{Il fu Mattia Pascal} - L. Pirandello; $p\!=\!0.20$, $\beta\!=\!1.07$
\item \textit{Uno, nessuno e centomila} - L. Pirandello; $p\!=\!0.16$, $\beta\!=\!1.05$
\item \textit{Il Corsaro Nero} - E. Salgari; $p\!=\!0.11$, $\beta\!=\!1.20$
\item \textit{Le tigri di Mompracem} - E. Salgari; $p\!=\!0.14$, $\beta\!=\!1.11$
\item \textit{Addio, amore!} - M. Serao; $p\!=\!0.14$, $\beta\!=\!1.26$
\item \textit{Il romanzo della fanciulla} - M. Serao; $p\!=\!0.12$, $\beta\!=\!1.31$
\item \textit{La coscienza di Zeno} - I. Svevo; $p\!=\!0.06$, $\beta\!=\!1.28$
\item \textit{Una vita} - I. Svevo; $p\!=\!0.06$, $\beta\!=\!1.26$
\item \textit{Fosca} - I. U. Tarchetti; $p\!=\!0.09$, $\beta\!=\!1.35$
\item \textit{Con gli occhi chiusi} - F. Tozzi; $p\!=\!0.10$, $\beta\!=\!1.35$
\item \textit{Tre croci} - F. Tozzi; $p\!=\!0.13$, $\beta\!=\!1.28$
\item \textit{I Malavoglia} - G. Verga; $p\!=\!0.04$, $\beta\!=\!1.62$
\item \textit{Mastro Don Gesualdo} - G. Verga; $p\!=\!0.13$, $\beta\!=\!1.27$
\end{packed_enum}

\vspace{0.5\baselineskip}
\noindent\underline{Spanish}

\begin{packed_enum}
\item \textit{La Regenta} - L. Alas; $p\!=\!0.11$, $\beta\!=\!1.21$
\item \textit{La vida de Lazarillo de Tormes} - Anonymous; $p\!=\!0.06$, $\beta\!=\!1.40$
\item \textit{La barraca} - V. Blasco Ibánez; $p\!=\!0.06$, $\beta\!=\!1.37$
\item \textit{Los cuatro jinetes del Apocalipsis} - V. Blasco Ibánez; $p\!=\!0.06$, $\beta\!=\!1.43$
\item \textit{Escenas andaluzas} - S. E. Calderón; $p\!=\!0.08$, $\beta\!=\!1.28$
\item \textit{Rayuela} - J. Cortázar; $p\!=\!0.12$, $\beta\!=\!1.12$
\item \textit{El sombrero de tres picos} - P. A. de Alarcón; $p\!=\!0.12$, $\beta\!=\!1.20$
\item \textit{Don Quijote de la Mancha} - M. de Cervantes; $p\!=\!0.10$, $\beta\!=\!1.22$
\item \textit{El Periquillo Sarniento} - J. J. F. de Lizardi; $p\!=\!0.07$, $\beta\!=\!1.33$
\item \textit{Penas arriba} - J. M. de Pereda; $p\!=\!0.06$, $\beta\!=\!1.32$
\item \textit{Sotileza} - J. M. de Pereda; $p\!=\!0.09$, $\beta\!=\!1.26$
\item \textit{La vida del Buscón} - F. de Quevedo; $p\!=\!0.09$, $\beta\!=\!1.30$
\item \textit{Abel Sánchez} - M. de Unamuno; $p\!=\!0.18$, $\beta\!=\!1.08$
\item \textit{Tirano Banderas} - R. M. del Valle-Inclán; $p\!=\!0.10$, $\beta\!=\!1.34$
\item \textit{Dona Bárbara} - R. Gallegos; $p\!=\!0.10$, $\beta\!=\!1.22$
\item \textit{Cien anos de soledad} - G. García Márquez; $p\!=\!0.04$, $\beta\!=\!1.42$
\item \textit{El Senor de Bembibre} - E. Gil y Carrasco; $p\!=\!0.07$, $\beta\!=\!1.26$
\item \textit{Sab} - G. Gómez de Avellaneda; $p\!=\!0.06$, $\beta\!=\!1.32$
\item \textit{El Criticón} - B. Gracián; $p\!=\!0.08$, $\beta\!=\!1.37$
\item \textit{Don Segundo Sombra} - R. Güiraldes; $p\!=\!0.07$, $\beta\!=\!1.43$
\item \textit{María} - J. Isaacs; $p\!=\!0.08$, $\beta\!=\!1.28$
\item \textit{A la costa} - L. A. Martínez; $p\!=\!0.08$, $\beta\!=\!1.27$
\item \textit{Amaya o los vascos en el siglo VIII} - F. Navarro Villoslada; $p\!=\!0.08$, $\beta\!=\!1.37$
\item \textit{Los pazos de Ulloa} - E. Pardo Bazán; $p\!=\!0.09$, $\beta\!=\!1.32$
\item \textit{Dona Perfecta} - B. Pérez Galdós; $p\!=\!0.10$, $\beta\!=\!1.23$
\item \textit{Fortunata y Jacinta} - B. Pérez Galdós; $p\!=\!0.08$, $\beta\!=\!1.31$
\item \textit{Misericordia} - B. Pérez Galdós; $p\!=\!0.10$, $\beta\!=\!1.31$
\item \textit{La vorágine} - J. E. Rivera; $p\!=\!0.09$, $\beta\!=\!1.33$
\item \textit{Pedro Páramo} - J. Rulfo; $p\!=\!0.11$, $\beta\!=\!1.25$
\item \textit{Facundo o civilización y barbarie en las pampas argentinas} - D. F. Sarmiento; $p\!=\!0.08$, $\beta\!=\!1.24$
\item \textit{Dona Luz} - J. Valera; $p\!=\!0.06$, $\beta\!=\!1.39$
\item \textit{Pepita Jiménez} - J. Valera; $p\!=\!0.06$, $\beta\!=\!1.41$
\end{packed_enum}

\vspace{0.5\baselineskip}
\noindent\underline{Polish}

\begin{packed_enum}
\item \textit{Ozimina} - W. Berent; $p\!=\!0.11$, $\beta\!=\!1.33$
\item \textit{Próchno} - W. Berent; $p\!=\!0.17$, $\beta\!=\!1.23$
\item \textit{Kariera Nikodema Dyzmy} - T. Dołęga-Mostowicz; $p\!=\!0.15$, $\beta\!=\!1.31$
\item \textit{As} - A. Dygański; $p\!=\!0.10$, $\beta\!=\!1.40$
\item \textit{Faraon} - A. Głowacki; $p\!=\!0.11$, $\beta\!=\!1.40$
\item \textit{Lalka} - A. Głowacki; $p\!=\!0.13$, $\beta\!=\!1.34$
\item \textit{Ferdydurke} - W. Gombrowicz; $p\!=\!0.20$, $\beta\!=\!1.15$
\item \textit{Trans-Atlantyk} - W. Gombrowicz; $p\!=\!0.19$, $\beta\!=\!1.22$
\item \textit{Inny świat} - G. Herling-Grudziński; $p\!=\!0.06$, $\beta\!=\!1.47$
\item \textit{Kamienie na szaniec} - A. Kamiński; $p\!=\!0.09$, $\beta\!=\!1.41$
\item \textit{Imperium} - R. Kapuściński; $p\!=\!0.12$, $\beta\!=\!1.38$
\item \textit{Król Maciuś Pierwszy} - J. Korczak; $p\!=\!0.08$, $\beta\!=\!1.68$
\item \textit{Stara baśń} - J. I. Kraszewski; $p\!=\!0.08$, $\beta\!=\!1.60$
\item \textit{Cudzoziemka} - M. Kuncewiczowa; $p\!=\!0.17$, $\beta\!=\!1.28$
\item \textit{Koroniarz w Galicji} - J. Lam; $p\!=\!0.09$, $\beta\!=\!1.41$
\item \textit{Zaklęty dwór} - W. Łoziński; $p\!=\!0.09$, $\beta\!=\!1.37$
\item \textit{Szaleństwa panny Ewy} - K. Makuszyński; $p\!=\!0.11$, $\beta\!=\!1.41$
\item \textit{Szatan z siódmej klasy} - K. Makuszyński; $p\!=\!0.11$, $\beta\!=\!1.42$
\item \textit{Trędowata} - H. Mniszek; $p\!=\!0.11$, $\beta\!=\!1.50$
\item \textit{Nad Niemnem} - E. Orzeszkowa; $p\!=\!0.13$, $\beta\!=\!1.24$
\item \textit{Chłopi} - W. Reymont; $p\!=\!0.10$, $\beta\!=\!1.46$
\item \textit{Ziemia obiecana} - W. Reymont; $p\!=\!0.11$, $\beta\!=\!1.40$
\item \textit{Dewajtis} - M. Rodziewiczówna; $p\!=\!0.15$, $\beta\!=\!1.39$
\item \textit{Lato leśnych ludzi} - M. Rodziewiczówna; $p\!=\!0.15$, $\beta\!=\!1.34$
\item \textit{Sklepy cynamonowe} - B. Schulz; $p\!=\!0.08$, $\beta\!=\!1.52$
\item \textit{Ogniem i mieczem} - H. Sienkiewicz; $p\!=\!0.10$, $\beta\!=\!1.42$
\item \textit{Quo vadis} - H. Sienkiewicz; $p\!=\!0.09$, $\beta\!=\!1.44$
\item \textit{W pustyni i w puszczy} - H. Sienkiewicz; $p\!=\!0.08$, $\beta\!=\!1.45$
\item \textit{Nienasycenie} - S. I. Witkiewicz; $p\!=\!0.10$, $\beta\!=\!1.38$
\item \textit{Ludzie bezdomni} - S. Żeromski; $p\!=\!0.11$, $\beta\!=\!1.37$
\item \textit{Popioły} - S. Żeromski; $p\!=\!0.11$, $\beta\!=\!1.37$
\item \textit{Przedwiośnie} - S. Żeromski; $p\!=\!0.13$, $\beta\!=\!1.29$
\item \textit{Poganka} - N. Żmichowska; $p\!=\!0.12$, $\beta\!=\!1.34$
\item \textit{Na srebrnym globie. Rękopis z Księżyca} - J. Żuławski; $p\!=\!0.08$, $\beta\!=\!1.51$
\end{packed_enum}

\vspace{0.5\baselineskip}
\noindent\underline{Russian}

\selectlanguage{russian}

\begin{packed_enum}
\item {\fontfamily{cmr} \foreignlanguage{russian}{\textit{Петербург (Peterburg)} - Б. Н. Бугаев (B. N. Bugaev);}} $p\!=\!0.23$, $\beta\!=\!1.13$
\item {\fontfamily{cmr} \foreignlanguage{russian}{\textit{Белая гвардия (Belaya gvardiya)} - М. А. Булгаков (M. A. Bulgakov);}} $p\!=\!0.22$, $\beta\!=\!1.15$
\item {\fontfamily{cmr} \foreignlanguage{russian}{\textit{Мастер и Маргарита (Master i Margarita)} - М. А. Булгаков (M. A. Bulgakov);}} $p\!=\!0.16$, $\beta\!=\!1.26$
\item {\fontfamily{cmr} \foreignlanguage{russian}{\textit{Театральный роман (Teatral'nyy roman)} - М. А. Булгаков (M. A. Bulgakov);}} $p\!=\!0.18$, $\beta\!=\!1.27$
\item {\fontfamily{cmr} \foreignlanguage{russian}{\textit{Деревня (Derévnya)} - И. А. Бунин (I. A. Bunin);}} $p\!=\!0.20$, $\beta\!=\!1.27$
\item {\fontfamily{cmr} \foreignlanguage{russian}{\textit{Жизнь Арсеньева (Zhizn' Arsen'yeva)} - И. А. Бунин (I. A. Bunin);}} $p\!=\!0.14$, $\beta\!=\!1.29$
\item {\fontfamily{cmr} \foreignlanguage{russian}{\textit{Мёртвые души (Mortvyye dushi)} - Н. В. Гоголь (N. V. Gogol');}} $p\!=\!0.16$, $\beta\!=\!1.28$
\item {\fontfamily{cmr} \foreignlanguage{russian}{\textit{Тарас Бульба (Taras bul'ba)} - Н. В. Гоголь (N. V. Gogol');}} $p\!=\!0.13$, $\beta\!=\!1.34$
\item {\fontfamily{cmr} \foreignlanguage{russian}{\textit{Обломов (Oblomov)} - И. А. Гончаров (I. A. Goncharov);}} $p\!=\!0.18$, $\beta\!=\!1.30$
\item {\fontfamily{cmr} \foreignlanguage{russian}{\textit{Обрыв (Obryv)} - И. А. Гончаров (I. A. Goncharov);}} $p\!=\!0.18$, $\beta\!=\!1.31$
\item {\fontfamily{cmr} \foreignlanguage{russian}{\textit{Бесы (Besy)} - Ф. М. Достоевский (F. M. Dostoyevskiy);}} $p\!=\!0.16$, $\beta\!=\!1.26$
\item {\fontfamily{cmr} \foreignlanguage{russian}{\textit{Братья Карамазовы (Brat'ya Karamazovy)} - Ф. М. Достоевский (F. M. Dostoyevskiy);}} $p\!=\!0.16$, $\beta\!=\!1.27$
\item {\fontfamily{cmr} \foreignlanguage{russian}{\textit{Преступление и наказание (Prestupleniye i nakazaniye)} - Ф. М. Достоевский (F. M. Dostoyevskiy);}} $p\!=\!0.19$, $\beta\!=\!1.22$
\item {\fontfamily{cmr} \foreignlanguage{russian}{\textit{Поединок (Poyedinok)} - А. И. Куприн (A. I. Kuprin);}} $p\!=\!0.21$, $\beta\!=\!1.16$
\item {\fontfamily{cmr} \foreignlanguage{russian}{\textit{Герой нашего времени (Geroy nashego vremeni)} - М. Ю. Лермонтов (M. Y. Lermontov);}} $p\!=\!0.15$, $\beta\!=\!1.36$
\item {\fontfamily{cmr} \foreignlanguage{russian}{\textit{Доктор Живаго (Doktor Zhivago)} - Б. Л. Пастернак (B. L. Pasternak);}} $p\!=\!0.14$, $\beta\!=\!1.30$
\item {\fontfamily{cmr} \foreignlanguage{russian}{\textit{Котлован (Kotlovan)} - А. П. Платонов (A. P. Platonov);}} $p\!=\!0.11$, $\beta\!=\!1.38$
\item {\fontfamily{cmr} \foreignlanguage{russian}{\textit{Чевенгур (Chevengur)} - А. П. Платонов (A. P. Platonov);}} $p\!=\!0.10$, $\beta\!=\!1.47$
\item {\fontfamily{cmr} \foreignlanguage{russian}{\textit{Капитанская дочка (Kapitanskaya dochka)} - А. С. Пушкин (A. S. Pushkin);}} $p\!=\!0.12$, $\beta\!=\!1.45$
\item {\fontfamily{cmr} \foreignlanguage{russian}{\textit{Путешествие из Петербурга в Москву (Puteshestviye iz Peterburga v Moskvu)} - А. Н. Радищев (A. N. Radishchev);}} $p\!=\!0.10$, $\beta\!=\!1.44$
\item {\fontfamily{cmr} \foreignlanguage{russian}{\textit{Архипелаг ГУЛАГ (Arkhipelag GULAG)} - А. И. Солженицын (A. I. Solzhenitsyn);}} $p\!=\!0.14$, $\beta\!=\!1.30$
\item {\fontfamily{cmr} \foreignlanguage{russian}{\textit{Пикник на обочине (Piknik na obochine)} - А. Н. \& Б. Н. Стругацкий (A. N. \& B. N. Strugatsky);}} $p\!=\!0.20$, $\beta\!=\!1.24$
\item {\fontfamily{cmr} \foreignlanguage{russian}{\textit{Анна Каренина (Anna Karenina)} - Л. Н. Толстой (L. N. Tolstoy);}} $p\!=\!0.14$, $\beta\!=\!1.34$
\item {\fontfamily{cmr} \foreignlanguage{russian}{\textit{Война и мир (Voyna i mir)} - Л. Н. Толстой (L. N. Tolstoy);}} $p\!=\!0.14$, $\beta\!=\!1.29$
\item {\fontfamily{cmr} \foreignlanguage{russian}{\textit{Воскресение (Voskreseniye)} - Л. Н. Толстой (L. N. Tolstoy);}} $p\!=\!0.14$, $\beta\!=\!1.30$
\item {\fontfamily{cmr} \foreignlanguage{russian}{\textit{Дворянское гнездо (Dvoryanskoye gnezdo)} - И. С. Тургенев (I. S. Turgenev);}} $p\!=\!0.14$, $\beta\!=\!1.39$
\item {\fontfamily{cmr} \foreignlanguage{russian}{\textit{Дым (Dym)} - И. С. Тургенев (I. S. Turgenev);}} $p\!=\!0.17$, $\beta\!=\!1.29$
\item {\fontfamily{cmr} \foreignlanguage{russian}{\textit{Новь (Nov')} - И. С. Тургенев (I. S. Turgenev);}} $p\!=\!0.18$, $\beta\!=\!1.30$
\item {\fontfamily{cmr} \foreignlanguage{russian}{\textit{Отцы и дети (Ottsy i deti)} - И. С. Тургенев (I. S. Turgenev);}} $p\!=\!0.16$, $\beta\!=\!1.33$
\item {\fontfamily{cmr} \foreignlanguage{russian}{\textit{Палата № 6 (Palata № 6)} - А. П. Чехов (A. P. Chekhov);}} $p\!=\!0.15$, $\beta\!=\!1.27$
\item {\fontfamily{cmr} \foreignlanguage{russian}{\textit{Три года (Tri goda)} - А. П. Чехов (A. P. Chekhov);}} $p\!=\!0.14$, $\beta\!=\!1.33$
\item {\fontfamily{cmr} \foreignlanguage{russian}{\textit{Тихий Дон (Tikhiy Don)} - М. А. Шолохов (M. A. Sholokhov);}} $p\!=\!0.19$, $\beta\!=\!1.26$
\end{packed_enum}

\selectlanguage{english}
\vspace{0.8cm}

\noindent
{\bf Translated texts}

\vspace{0.1cm}

\vspace{0.5\baselineskip}
\noindent\underline{English}

\vspace{0.5\baselineskip}
\begin{packed_itemize}
\item (1) \textit{David Copperfield}, $p\!=\!0.09$, $\beta\!=\!1.34$
\begin{packed_itemize}
\item German translation: $p\!=\!0.07$, $\beta\!=\!1.44$
\item French translation: $p\!=\!0.09$, $\beta\!=\!1.28$
\item Italian translation: $p\!=\!0.11$, $\beta\!=\!1.26$
\item Spanish translation: $p\!=\!0.10$, $\beta\!=\!1.18$
\item Polish translation: $p\!=\!0.15$, $\beta\!=\!1.27$
\item Russian translation: $p\!=\!0.12$, $\beta\!=\!1.37$
\end{packed_itemize}
\item (2) \textit{Treasure Island}, $p\!=\!0.12$, $\beta\!=\!1.18$
\begin{packed_itemize}
\item German translation: $p\!=\!0.07$, $\beta\!=\!1.50$
\item French translation: $p\!=\!0.11$, $\beta\!=\!1.26$
\item Italian translation: $p\!=\!0.12$, $\beta\!=\!1.19$
\item Spanish translation: $p\!=\!0.11$, $\beta\!=\!1.17$
\item Polish translation: $p\!=\!0.10$, $\beta\!=\!1.39$
\item Russian translation: $p\!=\!0.13$, $\beta\!=\!1.39$
\end{packed_itemize}
\end{packed_itemize}
\vspace{0.5\baselineskip}

\noindent\underline{German}

\vspace{0.5\baselineskip}
\begin{packed_itemize}
\item (3) \textit{Buddenbrooks}, $p\!=\!0.09$, $\beta\!=\!1.37$
\begin{packed_itemize}
\item English translation: $p\!=\!0.11$, $\beta\!=\!1.17$
\item Italian translation: $p\!=\!0.11$, $\beta\!=\!1.19$
\item Spanish translation: $p\!=\!0.11$, $\beta\!=\!1.12$
\item Polish translation: $p\!=\!0.11$, $\beta\!=\!1.35$
\item Russian translation: $p\!=\!0.15$, $\beta\!=\!1.25$
\end{packed_itemize}
\item (4) \textit{Die Leiden des jungen Werthers}, $p\!=\!0.07$, $\beta\!=\!1.57$
\begin{packed_itemize}
\item English translation: $p\!=\!0.07$, $\beta\!=\!1.41$
\item French translation: $p\!=\!0.09$, $\beta\!=\!1.35$
\item Italian translation: $p\!=\!0.09$, $\beta\!=\!1.25$
\item Spanish translation: $p\!=\!0.09$, $\beta\!=\!1.18$
\item Polish translation: $p\!=\!0.10$, $\beta\!=\!1.47$
\item Russian translation: $p\!=\!0.11$, $\beta\!=\!1.43$
\end{packed_itemize}
\end{packed_itemize}

\vspace{0.5\baselineskip}
\noindent\underline{French}

\vspace{0.5\baselineskip}
\begin{packed_itemize}
\item (5) \textit{Le Tour du monde en quatre-vingts jours}, $p\!=\!0.11$, $\beta\!=\!1.27$
\begin{packed_itemize}
\item English translation: $p\!=\!0.10$, $\beta\!=\!1.28$
\item German translation: $p\!=\!0.07$, $\beta\!=\!1.50$
\item Italian translation: $p\!=\!0.14$, $\beta\!=\!1.17$
\item Spanish translation: $p\!=\!0.10$, $\beta\!=\!1.28$
\item Polish translation: $p\!=\!0.10$, $\beta\!=\!1.40$
\item Russian translation: $p\!=\!0.12$, $\beta\!=\!1.32$
\end{packed_itemize}
\item (6) \textit{Les Misérables}, $p\!=\!0.10$, $\beta\!=\!1.25$
\begin{packed_itemize}
\item English translation: $p\!=\!0.09$, $\beta\!=\!1.33$
\item German translation: $p\!=\!0.06$, $\beta\!=\!1.53$
\item Italian translation: $p\!=\!0.12$, $\beta\!=\!1.23$
\item Spanish translation: $p\!=\!0.11$, $\beta\!=\!1.17$
\item Polish translation: $p\!=\!0.11$, $\beta\!=\!1.46$
\item Russian translation: $p\!=\!0.13$, $\beta\!=\!1.37$
\end{packed_itemize}
\end{packed_itemize}

\vspace{0.5\baselineskip}
\noindent\underline{Italian}

\vspace{0.5\baselineskip}
\begin{packed_itemize}
\item (7) \textit{Le avventure di Pinocchio}, $p\!=\!0.09$, $\beta\!=\!1.37$
\begin{packed_itemize}
\item English translation: $p\!=\!0.11$, $\beta\!=\!1.25$
\item German translation: $p\!=\!0.08$, $\beta\!=\!1.47$
\item French translation: $p\!=\!0.13$, $\beta\!=\!1.12$
\item Spanish translation: $p\!=\!0.13$, $\beta\!=\!1.16$
\item Polish translation: $p\!=\!0.11$, $\beta\!=\!1.44$
\item Russian translation: $p\!=\!0.14$, $\beta\!=\!1.33$
\end{packed_itemize}
\item (8) \textit{I promessi sposi}, $p\!=\!0.11$, $\beta\!=\!1.37$
\begin{packed_itemize}
\item English translation: $p\!=\!0.09$, $\beta\!=\!1.33$
\item German translation: $p\!=\!0.06$, $\beta\!=\!1.53$
\item French translation: $p\!=\!0.10$, $\beta\!=\!1.21$
\item Spanish translation: $p\!=\!0.11$, $\beta\!=\!1.33$
\item Polish translation: $p\!=\!0.11$, $\beta\!=\!1.37$
\item Russian translation: $p\!=\!0.13$, $\beta\!=\!1.35$
\end{packed_itemize}
\end{packed_itemize}

\vspace{0.5\baselineskip}
\noindent\underline{Spanish}

\vspace{0.5\baselineskip}
\begin{packed_itemize}
\item (9) \textit{Don Quijote de la Mancha}, $p\!=\!0.10$, $\beta\!=\!1.22$
\begin{packed_itemize}
\item English translation: $p\!=\!0.07$, $\beta\!=\!1.25$
\item German translation: $p\!=\!0.05$, $\beta\!=\!1.64$
\item French translation: $p\!=\!0.07$, $\beta\!=\!1.28$
\item Italian translation: $p\!=\!0.07$, $\beta\!=\!1.33$
\item Polish translation: $p\!=\!0.09$, $\beta\!=\!1.49$
\item Russian translation: $p\!=\!0.11$, $\beta\!=\!1.37$
\end{packed_itemize}
\item (10) \textit{La vida de Lazarillo de Tormes}, $p\!=\!0.06$, $\beta\!=\!1.40$
\begin{packed_itemize}
\item English translation: $p\!=\!0.05$, $\beta\!=\!1.35$
\item German translation: $p\!=\!0.05$, $\beta\!=\!1.67$
\item French translation: $p\!=\!0.09$, $\beta\!=\!1.25$
\item Italian translation: $p\!=\!0.07$, $\beta\!=\!1.25$
\item Polish translation: $p\!=\!0.08$, $\beta\!=\!1.51$
\item Russian translation: $p\!=\!0.10$, $\beta\!=\!1.44$
\end{packed_itemize}
\end{packed_itemize}
\vspace{0.5\baselineskip}

\noindent\underline{Polish}

\vspace{0.5\baselineskip}
\begin{packed_itemize}
\item (11) \textit{Quo vadis}, $p\!=\!0.09$, $\beta\!=\!1.44$
\begin{packed_itemize}
\item English translation: $p\!=\!0.08$, $\beta\!=\!1.33$
\item German translation: $p\!=\!0.06$, $\beta\!=\!1.57$
\item French translation: $p\!=\!0.11$, $\beta\!=\!1.23$
\item Italian translation: $p\!=\!0.09$, $\beta\!=\!1.25$
\item Spanish translation: $p\!=\!0.10$, $\beta\!=\!1.16$
\item Russian translation: $p\!=\!0.12$, $\beta\!=\!1.38$
\end{packed_itemize}
\item (12) \textit{Chłopi}, $p\!=\!0.10$, $\beta\!=\!1.46$
\begin{packed_itemize}
\item English translation: $p\!=\!0.08$, $\beta\!=\!1.42$
\item German translation: $p\!=\!0.05$, $\beta\!=\!1.61$
\item French translation: $p\!=\!0.07$, $\beta\!=\!1.38$
\item Spanish translation: $p\!=\!0.09$, $\beta\!=\!1.23$
\item Russian translation: $p\!=\!0.13$, $\beta\!=\!1.44$
\end{packed_itemize}
\end{packed_itemize}

\vspace{0.5\baselineskip}
\noindent\underline{Russian}
\vspace{0.5\baselineskip}

\begin{packed_itemize}
\item (13) {\fontfamily{cmr} \foreignlanguage{russian}{\textit{Братья Карамазовы (Brat'ya Karamazovy)}}}, $p\!=\!0.16$, $\beta\!=\!1.27$
\begin{packed_itemize}
\item English translation: $p\!=\!0.10$, $\beta\!=\!1.27$
\item German translation: $p\!=\!0.07$, $\beta\!=\!1.54$
\item French translation: $p\!=\!0.11$, $\beta\!=\!1.31$
\item Italian translation: $p\!=\!0.11$, $\beta\!=\!1.21$
\item Spanish translation: $p\!=\!0.10$, $\beta\!=\!1.27$
\item Polish translation: $p\!=\!0.13$, $\beta\!=\!1.39$
\end{packed_itemize}
\item (14) {\fontfamily{cmr} \foreignlanguage{russian}{\textit{Война и мир (Voyna i mir)}}}, $p\!=\!0.14$, $\beta\!=\!1.29$
\begin{packed_itemize}
\item English translation: $p\!=\!0.10$, $\beta\!=\!1.17$
\item German translation: $p\!=\!0.07$, $\beta\!=\!1.46$
\item French translation: $p\!=\!0.09$, $\beta\!=\!1.32$
\item Italian translation: $p\!=\!0.12$, $\beta\!=\!1.24$
\item Spanish translation: $p\!=\!0.11$, $\beta\!=\!1.18$
\item Polish translation: $p\!=\!0.06$, $\beta\!=\!1.69$
\end{packed_itemize}
\end{packed_itemize}

\end{document}